\documentclass[a4paper,fleqn]{cas-dc}
\usepackage[square, numbers, comma, sort&compress]{natbib}
\usepackage{setspace}
\usepackage{graphicx}
\usepackage{amsmath}
\usepackage{amssymb}
\usepackage{mathrsfs}
\usepackage{amsthm}
\usepackage{algorithm}
\usepackage{subcaption}
\usepackage{wrapfig, blindtext}
\usepackage[utf8x]{inputenc}
\usepackage[noend]{algpseudocode}
\usepackage{bm}
\usepackage{float}
\usepackage{todonotes}
\usepackage{parskip}
\newtheorem{assumption}{Assumption}
\newtheorem*{remark}{Remark}
\def\tsc#1{\csdef{#1}{\textsc{\lowercase{#1}}\xspace}}
\tsc{WGM}
\tsc{QE}
\tsc{EP}
\tsc{PMS}
\tsc{BEC}
\tsc{DE}

\DeclareMathOperator{\atantwo}{atan2}
\DeclareMathOperator{\clip}{clip}
\begin{document}
\let\WriteBookmarks\relax
\def\floatpagepagefraction{1}
\def\textpagefraction{.001}
\shorttitle{Taming an ASV for path following and collision avoidance using DRL}
\shortauthors{E Meyer et~al.}
\title [mode = title]{Taming an autonomous surface vehicle for path following and collision avoidance using deep reinforcement learning}                      
\author[1]{E Meyer}
\ead{eivind.meyer@gmail.com}
\author[1]{H Robinson}
\ead{haakon.robinson@ntnu.no}
\author[1,2]{A Rasheed}
\cormark[1]
\ead{adil.rasheed@ntnu.no}
\ead[url]{www.adilrasheed.com}
\address[1]{Norweigan University of Science and Technology, Elektro D/B2, 235, Gløshaugen, O. S. Bragstads plass 2, Trondheim, Norway}
\address[2]{Mathematics and Cybernetics, SINTEF Digital, Klæbuveien 153, Trondheim, Norway}
\author[3]{O San}
\ead{osan@okstate.edu}
\ead[url]{www.cfdlabs.org}
\address[3]{Oklahoma State University, 326 General Academic Building
Stillwater, Oklahoma 74078 USA}
\cortext[cor1]{Corresponding author}

\begin{abstract}
In this article, we explore the feasibility of applying  proximal policy optimization, a state-of-the-art deep reinforcement learning algorithm for continuous control tasks, on the dual-objective problem of controlling an underactuated autonomous surface vehicle to follow an a priori known path while avoiding collisions with non-moving obstacles along the way. The artificial intelligent agent, which is equipped with multiple rangefinder sensors for obstacle detection, is trained and evaluated in a challenging, stochastically generated simulation environment  based on the OpenAI gym python toolkit. Notably, the agent is provided with real-time insight into its own reward function, allowing it to dynamically adapt its guidance strategy. Depending on its strategy, which ranges from radical path-adherence to radical obstacle avoidance, the trained agent achieves an episodic success rate between 84 and 100\%
\end{abstract}
\begin{keywords}
Deep Reinforcement Learning \sep Autonomous Surface Vehicle \sep Collision Avoidance \sep Path Following \sep Machine Learning Controller
\end{keywords}
\maketitle
\section{Introduction}
Autonomy offers surface vehicles the opportunity to improve the efficiency of transportation while still cutting down on greenhouse emissions. However, for safe and reliable autonomous surface vehicles (ASV), effective path planning is a pre-requisite which should cater to the two important tasks of path following and collision avoidance (COLAV). In the literature, a distinction is typically made between \textit{reactive}  and \textit{deliberate} COLAV methods \cite{BeardMcLain}. In short, reactive approaches, most notably artificial potential field methods \cite{Khatib:1986:ROA:6806.6812, BorensteinKoren, Panagou}, dynamic window methods \cite{FoxDyn, highspeeddynwin, ModDynWid}, velocity obstacle methods \cite{journals/ijrr/FioriniS98, proactivecolavoidbrekke} and optimal control-based methods \cite{chenBarrier, hamiltonJacobiMITCHELL, branchingMPCEriksen, ibHagenMPC, BITAR2018389}, base their guidance decisions on sensor readings from the local environment, whereas deliberate methods, among them popular graph-search algorithms such as A* \cite{Hart1968} and Voronoi graphs \cite{Voronoi, fastmarching} as well as randomized approaches such as rapidly-exploring random tree \cite{LaValle1998RapidlyExploringRT} and probabilistic roadmap \cite{KavrakiProbRoadmap}, exploit a priori known characteristics of the global environment in order to construct an optimal path in advance, which is to be followed using a low-level steering controller. By utilizing more data than just the current perception of the local neighborhood surrounding the agent, deliberate methods are generally more likely to converge to the intended goal, and less likely to suggest guidance strategies leading to dead ends, which is frequently observed with reactive methods due to local minima \cite{Loe:Thesis}. However, in the case where the environment is not perfectly known, as a result of either incomplete or uncertain mapping data or due to the environment having dynamic features, purely deliberate methods often fall short. To prevent this, such methods are often executed repeatably on a regular basis to adapt to discrepancies between recent sensor observations and the a priori belief state of the environment \cite{Loe:Thesis}. However, as this class of methods are computationally expensive by virtue of processing global environment data, this is sometimes rendered infeasible for real-world applications with limited processing power \cite{WiigPhD}, especially as the problem of optimal path planning amid multiple obstacles is provably NP-hard \cite{Canny:1987:NLB:1382440.1383017}. Thus, a common approach is to utilize a reactive algorithm, which is activated whenever the presence of a nearby obstacle is detected, as a fallback option for the global, deliberate path planner. Such \textit{hybrid} architectures are intended to combine the strengths of reactive and deliberate approaches and have gained traction in recent years \cite{SERIGSTAD20181, eriksenHybridColreg}. The approach presented in this article is somewhat related to this; the existence of some a priori known nominal path is presumed, but following it strictly will invariantly lead to collisions with obstacles. Unlike other approaches, there is, however, no switching mechanism that activates some reactive fallback algorithm in dangerous situations. To this end, a reinforcement learning (RL) agent is trained to exhibit rational behaviour under such circumstances, i.e. following the path strictly only when it is deemed safe. RL is an area of machine learning (ML) of particular interest for control applications, such as the guidance of surface vessels under consideration here. Fundamentally, this ML paradigm is concerned with estimating the optimal behavior for an agent in an unknown, and potentially partly unobservable environment, relying on trial-and-error-like approaches in order to iteratively approximate the behavior policy that maximizes the agent's expected long-time reward in the environment. The field of RL has seen rapid development over the last few years, leading to many impressive achievements, such as playing chess and various other games at a level that is not only exceedingly superhuman, but also overshadows previous AI approaches by a wide margin \cite{alphazeropaper, alphago, alphastarblog}. 

The focus of this paper is to explore how RL, given the recent advances in the field, can be applied to the guidance and control of ASV. Specifically, we look at the dual objective of achieving the ability to follow a path constructed from a priori known way-points, while avoiding colliding into obstacles along the way. For the purpose of simplicity, we limit the scope of this work to non-moving obstacles of circular shapes. As RL methods are, by their very nature, model-free approaches, a positive result can bring significant value to the robotics field, where implementing a guidance system typically requires knowledge of the vessel dynamics, which often relies on non-linear first-principles models with parameters that can only be determined experimentally.

\section{Theory}
\label{section:theory}
\subsection{Guidance and control of marine vessels}
\subsubsection{Coordinate frames}

In order to model the dynamics of marine vessels, one must first define the coordinate frames forming the basis for the motion. A few coordinate frames typically used in control theory are of particular interest. The geographical North-East-Down (NED) reference frame $\{n\} = (x_n, y_n, z_n)$ forms a tangent plane to the Earth's surface, making it useful for terrestrial navigation. Here, the $x_n$-axis is directed north, the $y_n$-axis is directed east and the $z_n$-axis is directed towards the center of the earth.

The origin of the body-fixed reference frame $\{b\} = (x_b, y_b, z_b)$ is fixed to the current position of the vessel in the NED-frame, and its axes are aligned with the heading of the vessel such that $x_b$ is the longitudinal axis, $y_b$ is the transversal axis and $z_b$ is the normal axis pointing downwards. It should be noted, that whenever the vessel is aligned with the water surface, which is an assumption that is typically made, $z_b$ points in the same direction as $z_n$, i.e. towards the center of the Earth.

\subsubsection{State variables}\label{section:sname_state_variables}

Following Society of Naval Architects and Marine Engineers (SNAME) notation \cite{SNAME}, twelve variables are used for representing the vessel state. The state vector consists of the generalized coordinates $\bm{\eta} \triangleq \left[ x^n, y^n, z^n, \phi, \theta, \psi \right]^T$, where the quantities in the bracket are North, East, Down positions in reference frame $\{n\}$, roll, pitch, yaw corresponding to a Euler angle \textit{zyx} convention from $\{n\}$ to $\{b\}$  respectively, representing the pose of the vessel relative to the inertial frame.
Also $\bm{\nu} \triangleq \left[ u, v, w, p, q, r \right]^T$, where the quantities in the bracket are surge, sway, heave, roll rate, pitch rate and yaw rate respectively representing the vessel's translational and angular velocity in the body-frame.




\subsubsection{Dynamics}\label{section:vessel_dynamics}

\begin{assumption}[Calm sea]\label{as:calm_sea} There is no ocean current, no wind and no waves and thus no external disturbances to the vessel.\end{assumption}

\noindent In the general case, twelve coupled, first-order, nonlinear ordinary differential equations make up the vessel dynamics. In the absence of ocean currents, waves and wind, these can be expressed in a compact matrix-vector form as 
\begin{equation}
\begin{aligned}
\dot{\bm{\eta}} &= \mathbf{J}_{\bm{\Theta}}(\bm{\eta})   \bm{\nu} \\
\mathbf{B} \bm{f} &= \mathbf{M_{RB}} \bm{\dot{\nu}} + \mathbf{C_{RB}}(\bm{\nu}) \bm{\nu} + \bm{g}(\bm{\eta}) &&\text{\scriptsize (rigid-body, hydrostatic)}\\
&+ \mathbf{M_A} \bm{\dot{\nu}} + \mathbf{C_{A}}(\bm{\nu}) \bm{\nu} + \mathbf{D}(\bm{\nu}) &&\text{\scriptsize (hydrodynamic)} 
\end{aligned}
\end{equation}
%
%
\noindent Here, $\mathbf{J}_{\bm{\Theta}}(\bm{\eta})$ is the transformation matrix from the body frame $\{b\}$ to the NED reference frame $\{n\}$. $\mathbf{M_{RB}}$ and $\mathbf{M_A}$ are the mass matrices representing rigid-body mass and added mass, respectively. Analogously, $\mathbf{C_{RB}}(\bm{\nu})$ and $\mathbf{C_{A}}(\bm{\nu})$ are matrices incorporating centripetal and Coriolis effects. Finally, $\mathbf{D}(\bm{\nu})$ is the damping matrix, $\bm{g}(\bm{\eta})$ contains the restoring forces and moments resulting from gravity and buoyancy, $\mathbf{B}$ is the actuator configuration matrix and $f$ is the vector of control inputs.

\subsubsection{3-DOF maneuvering model}\label{section:vessel_model_3dof}
In this subsection, the ASV assumptions and the resulting 3-DOF model is outlined.

\begin{assumption}[State space restriction]\label{as:restricted_motion}The vessel is always located on the surface and thus there is no heave motion. Also, there is no pitching or rolling motion.\end{assumption}

\noindent This assumption implies that the state variables $z^{n}$, $\phi$, $\theta$, $w$, $p$, $q$ are all zero. Thus, we are left with the three generalized coordinates $x^n$, $y^n$ and $\psi$ and the body-frame velocities $u$, $v$ and $r$. In this case, the transformation matrix $\mathbf{J}_{\bm{\Theta}}(\bm{\eta})$ is reduced to a basic rotation matrix $\mathbf{R}_{z,\psi}$ for a rotation of $\psi$ around the $z_n$-axis as defined by
\begin{equation*}
\begin{aligned}
\mathbf{R}_{z,\psi} &= \begin{bmatrix}
    \cos{\psi} & -\sin{\psi} & 0 \\
    \sin{\psi} & \cos{\psi} & 0 \\
    0 & 0 & 1
\end{bmatrix}
\end{aligned}
\end{equation*}
Furthermore, as vertical motion is disregarded, we have that $\bm{g}(\bm{\eta}) = \bm{0}$. Also, by combining the corresponding rigid-body and added mass terms associated such that $\mathbf{M} = \mathbf{M_{RB}} + \mathbf{M_B}$ and $\mathbf{C}(\bm{\nu}) = \mathbf{C_{RB}}(\bm{\nu}) + \mathbf{C_{A}}(\bm{\nu})$, we obtain the simpler 3-DOF state-space model
\begin{equation}
\begin{aligned}
\dot{\bm{\eta}} &= \mathbf{R}_{z,\psi}(\bm{\eta})   \bm{\nu} \\
\mathbf{M} \bm{\dot{\nu}} + \mathbf{C}(\bm{\nu}) \bm{\nu} + \mathbf{D}(\bm{\nu}) &= \mathbf{B} \bm{f}
\end{aligned}
\end{equation}

\noindent where $\bm{\eta} \triangleq \left[x^n, y^n, \psi \right]^T$ and $\bm{\nu} \triangleq \left[u, v, r \right]^T$ and each matrix is $3$x$3$. 

\begin{assumption}[Vessel symmetry]\label{as:vessel_symmetry}The vessel is port-starboard symmetric.\end{assumption}

\begin{assumption}[Origin at the centerline]\label{as:origin_centerline}The body-fixed reference frame $\{b\}$ is centered somewhere at the longitudinal centerline passing through the vessel's center of gravity.\end{assumption}

\begin{assumption}[Sway-underactuation]\label{as:underactuation}There is no force input in sway, so the only control inputs are the propeller thrust $T_u$ and the rudder angle $T_r$.\end{assumption}

\begin{assumption}[Linear-quadratic damping]\label{as:linquaddamping}The damping model includes linear and quadratic effects.\end{assumption}

\noindent Assumptions \ref{as:vessel_symmetry} and \ref{as:origin_centerline}, which are commonly found in maneuvering theory applications, justify a sparser structure of the system matrices, where some non-diagonal elements are zeroed out. Also, from Assumption \ref{as:underactuation} we have that $\bm{f} \triangleq \left[T_u, T_r\right]^T$. The matrices and numerical values are obtained from a 3-DOF adaptation of \cite{daSilvaModel}, where they were estimated partly through field experiments for a 6-DOF torpedo-shaped submarine.
%

\subsection{Reinforcement Learning}
In this section, we will briefly review the RL paradigm and introduce the specific technique that our method builds on. For a more comprehensive coverage, the reader is advised to consult the book by Sutton and Barto \cite{Sutton1998}.

Fundamentally, RL is an approach to let autonomous agents learn how to behave optimally in their environments. Using the phrase "let learn" instead of "teach" is not accidental; a defining feature of RL is that the learning is not instructive, as opposed to the related field of supervised learning. In other words, training the agent through training samples defining the optimal action under the given conditions is not RL. Instead, learning is achieved through a combination of exploration and evaluative feedback, which bears a close resemblance to the way in which humans and other animals learn \cite{Sutton1998}; they become gradually wiser by virtue of trial and error. 

\subsubsection{Fundamentals of RL}\label{section:fundamental_RL}
At each discrete time-step of the learning process, the \textbf{agent}, which is operating within an \textbf{environment}, chooses an \textbf{action} $u$ based on its current \textbf{state} $s$ (also often referred to as \textit{observation}). The way in which the specific action was chosen by the agent (i.e. the agent's strategy) is commonly referred to as the \textbf{policy} and denoted by $\pi$. Thus, the policy $\pi$ can be thought of as a mapping $\pi:\mathcal S\to\mathcal A$ from the state space to the action space. In order to learn, i.e. improve the policy $\pi$, the agent then receives a numerical \textbf{reward} $r$ from the environment. The fundamental goal of the agent is to maximize its long-term reward (also known as the \textbf{return}), and updates to the agent's policy are intended to improve the agent's ability to do this. These concepts (i.e. \textit{agents}, \textit{environments}, \textit{observations/states}, \textit{policies}, \textit{actions} and \textit{rewards}) are fundamental to the study of RL.
\begin{remark}
The reward may not solely depend on the latest action made. An intuitively attractive action may have long-term repercussions. Similarly, an action which is unexciting in the short-term may be optimal in the long term. Delayed rewards are common in RL environments.
\end{remark}
\begin{remark}
The policy need not be deterministic. In fact, in games such as rock–paper–scissors, the optimal policy is stochastic.
\end{remark}
\begin{remark}
The actions need not be discrete. Traditionally, RL algorithm have been dealing with discrete action spaces, but recent advances in the field have led to state-of-the-art algorithms that are naturally compatible with continuous action spaces (i.e. do not involve the workaround of discretizing a continuous action space, which is undesirable for control applications \cite{lillicrap2015continuous}). 
\end{remark}


\noindent As the environment may be stochastic, it is common to think of the process as a Markov decision process (MDP) with state space $\mathcal{S}$, action space $\mathcal{A}$, reward function $r(s_t, a_t)$, transition dynamics $p(s_{t+1}| s_t, a_t)$ and an initial state distribution $p(s_0)$ \cite{Bertsekas:2000:DPO:517430}. The combined MDP and agent formulation allows us to sample trajectories from the process by first sampling an initial state from $p(s_0)$, and then repeatedly sampling the agent's action ${a_t}$ from its policy $\pi(s_t)$ and the next state $s_{t+1}$ from $p(s_{t+1}| s_t, a_t)$. As the agent is rewarded at each time step, its total reward can be represented as
\begin{equation}
R_t \triangleq \sum_{i=t}^{\infty}{r(s_i, a_i)}
\end{equation}
\begin{remark}
Analogous to discount functions used in the field of economics, it is common to introduce a discount factor $\gamma \in (0, 1]$ to capture the agent's relative preference for short-term rewards mathematically and to ensure that the infinite sum of rewards will not diverge. The discounted sum of rewards is then given by $\sum_{t=0}^{\infty}{\gamma^t r(s_t, a_t)}$. For concreteness in the following derivations, however, the discount factor is disregarded. This is justified by considering the discount factor as being already incorporated into the reward function, making it time-dependent.
\end{remark}

\noindent Due to the stochasticity of the environment, one must consider the expected sum of rewards to obtain a tractable formulation for optimization purposes. Thus, we can introduce the state-value function $V^{\pi}(s)$ and the action-value function $Q^{\pi}(s, a)$, two very related concepts. $V^{\pi}(s)$ represents the expected return from time $t$ onwards given an initial state $s$, whereas $Q^{\pi}(s, a)$ represents the expected return from time $t$ onwards \textbf{conditioned on the initial action $a_t$}. 
\begin{align}
V^{\pi}(s_t) &\triangleq \mathbb{E}_{s_{i>=t}, a_{i>=t} \sim \pi} \left[ R_t | s_t \right]\\
Q^{\pi}(s_t, a_t) &\triangleq \mathbb{E}_{s_{i>=t}, a_{i>=t} \sim \pi} \left[ R_t | s_t, a_t \right]
\end{align}
\subsubsection{Policy gradients}

Whereas value-based methods are concerned with estimating the state-value function and then inferring the optimal policy, policy-based methods directly optimize the policy. For high-dimensional or continuous action spaces, policy-based methods are commonly considered to be the more efficient approach \cite{tai2016survey}. 

From now on, we consider the policy $\pi(\theta)$ to be stochastic (i.e. $\pi(\theta):\mathcal S \times \mathcal{A} \to [0, 1]$) and assume that is defined by some differentiable function parametrized by $\theta$, enabling us to optimize it through policy-gradient methods. In general, these methods are concerned with using gradient ascent approximations to gradually adjust the policy function parameterization vector in order to optimize the performance objective
\begin{equation}
J(\theta) \triangleq \mathbb{E}_{s_i, a_i \sim \pi(\bm{\theta)}} \left[ R_0 \right]
\end{equation}
More formally, policy-gradient methods approach gradient ascent by updating the parameter vector $\theta$ according to the approximation $ \theta_{t+1} \gets \alpha \theta_{t} + \widehat{\nabla_{\theta} J(\theta)}$, where $\widehat{\nabla_{\theta} J(\theta)}$ is a stochastic estimate of $\nabla_{\theta} J(\theta)$ satisfying $\mathbb{E}{\left[ \widehat{\nabla_{\theta} J(\theta)} \right]} = \nabla_{\theta} J(\theta)$. Intuitively, the estimation of the policy gradient might be considered intractible, as the state transition dynamics, which affect the expected reward and hence our performance objective, are influenced by the agent's policy in an unknown fashion. However, the policy gradient theorem \cite{pgmforreinforcementlearning} establishes that the policy gradient $\nabla_{\theta} J(\theta)$ satisfies
\begin{equation}\label{eq:policy_grad_theorem}
\nabla_{\theta} J(\theta) \propto \sum_{s}{\mu(s)\sum_{a}{\nabla_{\theta} \pi(a | s) Q^{\pi}(s, a)} }   
\end{equation}
\noindent Here, $\mu$ is the steady state distribution under $\pi$, i.e. $\mu(s) = \lim_{t\to\infty}{Pr\{S_t = s | A_{0:{t-1}} \sim \pi\}}$, where $S_t$ and $A_{0:{t-1}}$ are random variables representing the state at time-step $t$, and the actions up to that point, respectively. Interestingly, the expression for the policy gradient does not contain the derivative $\nabla_{\theta} \mu(s)$, implying that approximating the gradient by sampling is feasible, because calculating the effect of updating the policy on the steady state distribution is not needed. By replacing the probability-weighted sum over all possible states in Equation \ref{eq:policy_grad_theorem} by an expectation of the random variable $S_t$ under the current policy, we have that
%
\begin{equation}
\nabla_{\theta} J(\theta) \propto \mathbb{E}_{\pi}{ \left[ \sum_{a}{\nabla_{\theta} \pi(a | S_t) Q^{\pi}(S_t, a)} \right] }    
\end{equation}
\noindent Similarly, we can replace the sum over all possible actions with an expectation of the random variable $A_t$ after multiplying and dividing by the policy $\pi(a | S_t)$:
%
\begin{equation*}
\nabla_{\theta} J(\theta) \propto \mathbb{E}_{\pi}{ \left[ \sum_{a}{\frac{\pi(a | S_t)}{\pi(a | S_t)} \nabla_{\theta} \pi(a | S_t) Q^{\pi}(S_t, a)} \right] }    
\end{equation*}
%
\begin{equation}
\nabla_{\theta} J(\theta) \propto \mathbb{E}_{\pi}{\left[ \frac{\nabla_{\theta} \pi(A_t | S_t)}{\pi(A_t | S_t)} Q^{\pi}(S_t, A_t) \right] }   
\end{equation}
\noindent Furthermore, it follows from the identity $\nabla{\ln{x}} = \frac{\nabla{x}}{x}$ that
%
\begin{equation}
\nabla_{\theta} J(\theta) \propto \mathbb{E}_{\pi}{\left[ \nabla_{\theta}{\ln{\pi(A_t | S_t)}} Q^{\pi}(S_t, A_t) \right]}   
\end{equation}
\noindent Also, by considering that
\begin{equation}
\begin{aligned}
\sum\limits_{a}{b(s) \nabla{\pi(a|s)}} &= b(s) \nabla{\sum\limits_{a}{\pi(a|s)}} \\
&= b(s) \nabla{\bm{1}} = 0
\end{aligned}
\end{equation}
\noindent it is straight-forward to see that one can replace the state-action value function $Q^{\pi}(s, a)$ in Equation \ref{eq:policy_grad_theorem} by $Q^{\pi}(s, a) - b(s)$, where the \textbf{baseline} function $b(s)$ can be an arbitrary function not depending on the action $a$, without introducing a bias in the estimate. However, it can be shown that the variance of the estimator can be greatly reduced by introduction such a baseline. It is possible to calculate the optimal (i.e. variance-minimizing) baseline \cite{weaver2013optimal}, but commonly the state value function $V^{\pi}$ is used, yielding an almost optimal variance \cite{schulmanhighdimcontrol}. The resulting term, is known as the advantage function:
\begin{equation}\label{eq:advantage_function}
A^{\pi}(s, a) = Q^{\pi}(s, a) - V^{\pi}(s)
\end{equation}
\noindent which intuitively represents the expected improvement obtained by an action compared to the default behavior. Furthermore, by following the same steps as outlined above, we end up with the expression
%
\begin{equation}
\nabla_{\theta} J(\theta) \propto \mathbb{E}_{\pi}{\left[ \nabla_{\theta}{\log{\pi(A_t | S_t)}} A^{\pi}(s, a) \right]}   
\end{equation}
\noindent Thus, an unbiased empirical estimate based on $N$ episodic trajectories (i.e. independent rollouts of the policy in the environment) of the policy gradient is
\begin{equation}\label{eq:reinforce_emp_grad}
    \widehat{\nabla_{\theta} J(\theta)} = \frac{1}{N} \sum_{n=1}^{N}{\sum_{t=0}^{\infty}{\hat{A}_{t}^{n} \nabla_{\theta} \log{\pi(a_t^n | s_t^n})   }}
\end{equation}

\subsubsection{Advantage function estimation}\label{section:gae}
As both $Q^{\pi}(s, a)$ and $V^{\pi}(s)$ are unknown in general, it follows that $A^{\pi}(s, a)$ is also unknown. Thus, it is commonly replaced by an advantage estimator $\hat{A}^{\pi}(s, a)$. Various estimation methods have been developed for this purpose, but a particularly popular one is Generalized Advantage Estimation (GAE) as originally outlined in \cite{schulmanhighdimcontrol}, which uses discounted temporal difference (TD) residuals of the state value function as the fundamental building blocks. For this, we reintroduce the discount parameter $\gamma$. However, even if $\gamma$ corresponds to the discount factor discussed in the context of MDPs, we now consider it as a variance-reducing parameter in an undiscounted MDP. TD residuals \cite{Sutton1998}, which are in widespread use within RL, and give a basic estimate of the advantage function, are defined by
\begin{equation}
    \delta_t^V = r_t + \gamma \hat{V}(s_{t+1}) - \hat{V}(s_t)
\end{equation}

\noindent where $\hat{V}$ is an approximate value function. Whenever $\hat{V} = V^\pi$, i.e. our approximation equals the real value function, the estimate is actually unbiased. For practical purposes, however, this is unlikely to be the case, so a common approach is to look further ahead than just one step in order to reduce the bias. More formally, by defining $\hat{A}_t^{(k)}$ as the discounted sum of the $k$ next TD residuals, we have that
\begin{equation}
\begin{aligned}
    \hat{A}_t^{(1)} &= \delta_t^{\hat{V}} = - \hat{V}(s_t) + r_t + \gamma \hat{V}(s_{t+1}) \\
    \hat{A}_t^{(2)} &= \delta_t^{\hat{V}} + \gamma \delta_{t+1}^{\hat{V}} = - \hat{V}(s_t) + r_t + \gamma r_{t+1} + \gamma^2 \hat{V}(s_{t+2})  \\
    \vdots \\
    \hat{A}_t^{(k)} &= \sum_{l=0}^{k-1} \gamma^{l}\delta_{t+l}^{\hat{V}}
\end{aligned}
\end{equation}
\noindent The defining feature of GAE is that, instead of choosing some k-step estimator $\hat{A}_t^{(k)}$, we use an exponentially weighted average of the $k$ first estimators, letting $k \to \infty$. Thus, we have that
\begin{equation}
\begin{aligned}
    \hat{A}_t^{GAE(\gamma, \lambda)} \triangleq (1 - \lambda) (\hat{A}_t^{1} + \lambda \hat{A}_t^{2} + \lambda^2 \hat{A}_t^{3} + \dots)
\end{aligned}
\end{equation}
\noindent which can be shown by insertion of the definition of $\hat{A}_t^{(k)}$ to equal
\begin{equation}
    \hat{A}_t^{GAE(\gamma, \lambda)} = \sum_{l=0}^{\infty}{(\gamma \lambda)}^l \delta_{t+l}^{\hat{V}}
\end{equation}

\noindent Here, $\lambda \in [0, 1]$ serves as a trade-off parameter controlling the compromise between bias and variance in the advantage estimate; using a small value lowers the variance as the immediate TD residuals make up most of the estimate, whereas using a large value lowers the bias induced from inaccuracies in the value function approximation.

Due to the recent advances made within deep learning (DL), a common approach is to use a deep neural network (DNN) for estimating the value function, which is trained on the discounted empirical returns. More specifically, the DNN state value estimator $\hat{V}_\theta(s_t)$, which is parametrized by $\theta_{VF}$, is trained by minimizing the loss function
\begin{equation}
    L_t^{VF}(\theta) = \hat{\mathbb{E}}_t {\left[ \hat{V}_\theta(s_t) - \sum_{i=t}^{\infty}{\gamma^{i-t} r(s_i, a_i)} \right]}
\end{equation}

\noindent where the expectation $\hat{\mathbb{E}}_t{\left[...\right]}$ represents the empirical average obtained from a finite batch of samples. The reader is referred to \cite{goodfellowDL} for a comprehensive introduction to DL, or to \cite{Bishop:2006:PRM:1162264}, which covers supervised machine learning, of which DL is a subfield.

\subsubsection{A surrogate objective}\label{section:surrogate_objective}

Optimizing the performance objective directly using the empirical policy gradient approximation from Equation \ref{eq:reinforce_emp_grad} is feasible; in fact, this constitutes the vanilla policy gradient algorithm originally proposed in \cite{Williams:1992:SSG:139611.139614}. However, it is well known that this approach has limitations due to a relatively low sample efficiency and thus suffers from a rather slow convergence time, as it requires an excessive number of samples for accurately estimating the policy gradient direction \cite{Kakade:2002:AOA:645531.656005}. Accordingly, unless the step-size is chosen to be trivially small (yielding unacceptably slow convergence), it is not guaranteed that the policy update will improve the performance objective, which leads to the algorithm having poor stability and robustness characteristics. \cite{Schulman2016OptimizingEF}. 

Instead, recent state-of-the-art policy gradient methods such as Trust Region Policy Optimization (TRPO) \cite{SchulmanLMJA15_TRPO} and its "successor" Proximal Policy Optimization \cite{schulman2017proximal} optimize a surrogate objective function which provides theoretical guarantees for policy improvement even under nontrivial step sizes. Fundamentally, these methods rely on the relative policy performance identity proven in \cite{Kakade:2002:AOA:645531.656005}, which states that the improvement in the performance objective $J(\theta)$ achieved by a policy update $\theta \to \theta'$ is equal to the expected advantage (ref. Equation \ref{eq:advantage_function}) of the actions sampled from the new policy $\pi_{\theta'}'$ calculated with respect to the old policy $\pi_{\theta}$. More formally, this translates to
\begin{equation}\label{eq:relative_policy_performance_identity}
    J(\theta') - J(\theta) = \mathbb{E}_{\pi'_{\theta}}{ \left[ \sum_{t}{\gamma^t A^{\pi_{\theta}} (s_t, a_t)} \right]}
\end{equation}

\noindent which is, albeit interesting, not practically useful as the expectation is defined under the next (i.e. unknown) policy $\pi_{\theta'}$, which we are obviously unable to sample trajectories from. However, Equation \ref{eq:relative_policy_performance_identity} can be rewritten and finally approximated by
\begin{equation}
\begin{aligned}
  &J(\theta') - J(\theta) \\&= \textstyle\sum_{t}{\mathbb{E}_{s_t \sim \pi_{\theta'}}{\left[\mathbb{E}_{a_t \sim \pi_{\theta'}}{\left[ \gamma^t A^{\pi_{\theta}} (s_t, a_t) \right]} \right]}} \\
   &= \textstyle\sum_{t}{\mathbb{E}_{s_t \sim \pi_{\theta'}}{\left[\mathbb{E}_{a_t \sim \pi_{\theta}}{\left[\tfrac{\pi_{\theta'}{(a_t | s_t)}}{\pi_{\theta}{(a_t | s_t)}} \gamma^t A^{\pi_{\theta}} (s_t, a_t) \right]} \right]}}\\
   &\approx \textstyle\sum_{t}{\mathbb{E}_{s_t \sim \pi_{\theta}}{\left[\mathbb{E}_{a_t \sim \pi_{\theta}}{\left[\tfrac{\pi_{\theta'}{(a_t | s_t)}}{\pi_{\theta}{(a_t | s_t)}} \gamma^t A^{\pi_{\theta}} (s_t, a_t) \right]} \right]}}
\end{aligned}
\end{equation}

\noindent Where the third and last steps can be seen as \textit{importance sampling} and neglecting \textit{state distribution mismatch respectively}. Loosely stated, the last approximation assumes that the change in the state distribution induced by a small update to the policy parameters is negligible. This is justified by theoretical guarantees imposing an upper bound to the distribution chance provided in \cite{Kakade:2002:AOA:645531.656005}. This suggests that one can reliably optimize the \textit{conservative policy iteration} surrogate objective
\begin{equation}
    J^{CPI}(\theta') = \hat{\mathbb{E}}_t {\left[ \frac{\pi_{\theta'}{(a_t | s_t)}}{\pi_{\theta}{(a_t | s_t)}} \hat{A}_t^{\pi_{\theta}}  \right]}
\end{equation}

\noindent \cite{Kakade:2002:AOA:645531.656005}. However, this approximation is only valid in a local neighborhood, requiring a carefully chosen step-size to avoid instability. In TRPO, this is achieved by maximizing $L^{CPI}(\theta')$ under a hard constraint on the KL divergence between the old and the new policy. However, as this is computationally expensive, the PPO algorithm refines this by integrating the constraint into the objective function by redefining the objective function to
\begin{equation}
\begin{aligned}
    &J^{CLIP}(\theta^{\prime}) = \hat{\mathbb{E}}_t {\left[\min{\left( r_t(\theta) \hat{A}_t^{\pi_{\theta}}, \clip_\epsilon{(r_t(\theta))} \hat{A}_t^{\pi_{\theta}} \right)}  \right]}\\
    &\clip_\epsilon(x) = \clip{(x, 1 - \epsilon, 1 + \epsilon)}
\end{aligned}
\end{equation}

\noindent where $r_t(\theta)$ is a shorthand notation for the probability ratio $\frac{\pi_{\theta'}{(a_t | s_t)}}{\pi_{\theta}{(a_t | s_t)}}$. The truncation of the probability ratio is motivated by a need to restrict $r_t(\theta)$ from moving outside of the interval $\left[1 - \epsilon, 1 + \epsilon\right]$. Also, the expectation is taken over the minimum of the clipped and unclipped objective, implying that the overall objective function is a lower bound of the original objective function $J^{CPI}(\theta')$. At each training iteration, the advantage estimates are computed over batches of trajectories collected from $N_A$ concurrent actors, each of which executes the current policy $\pi_{\theta}$ for $T$ timesteps. Afterwards, a stochastic gradient descent (SGD) update using the \textit{Adam} optimizer \cite{kingma2014adam} of minibatch size $N_{MB}$ is performed for $N_E$ epochs.

\begin{algorithm}[!htb]
\scriptsize
    \begin{algorithmic}
    \For{iteration = $1, 2, ...$}
    \For{actor = $1, 2, ... N$}
        \State For $T$ time-steps, execute policy $\pi_{\theta}$.
        \State Compute advantage estimates $\hat{A}_1, \dots \hat{A}_T$
    \EndFor
    \For{epoch = $1, 2, ... N_E$}
        \State Obtain mini batch of $N_{MB}$ samples from the $N_A T$ simulated time-steps.
        \State Perform SGD update from minibatch $(\mathbf{X}_{MB}, \mathbf{Y}_{MB})$.
        \State $\theta \gets \theta'$
    \EndFor
    \EndFor
    \end{algorithmic}
\caption{Proximal Policy Optimisation}
\label{alg:ppo}
\end{algorithm}

The PPO algorithms strikes a balance between ease of implementation and data
efficiency, and, unlike former state-of-the-art methods, is likely to perform well in a wide range of continuous environments without extensive hyperparameter tuning \cite{schulman2017proximal}.

\subsection{Tools and libraries}

The code implementation of our solution make use of the RL framework provided by the Python library \textbf{OpenAI Gym} \cite{OPENAIGYM}, which was created for the purpose of standardizing the benchmarks used in RL research. It provides a easy-to-use framework for creating RL environments in which custom RL agents can be deployed and trained with minimal overhead. 

\textbf{Stable Baselines} \cite{stable-baselines}, another Python package, provides a large set of state-of-the-art parallelizable RL algorithms compatible with the OpenAI gym framework, including PPO. The algorithms are based on the original versions found in OpenAI Baselines \cite{OPENAIBASELINES}, but Stable Baselines provides several improvements, including algorithm standardization and exhaustive documentation.

\section{Methodology}
In this section, we outline the specifics of our approach by defining the fundamental RL concepts as presented in Section \ref{section:fundamental_RL} according to the problem at hand and describe how the vessel's guidance capabilities are trained within the context of the RL framework Stable Baselines.

\subsection{Environment}
\begin{figure}[pos=!htb]
\centering
\begin{subfigure}[]{0.49\linewidth}
    {\includegraphics[trim={2cm 0.7cm 3cm 1cm},clip,width=\linewidth]{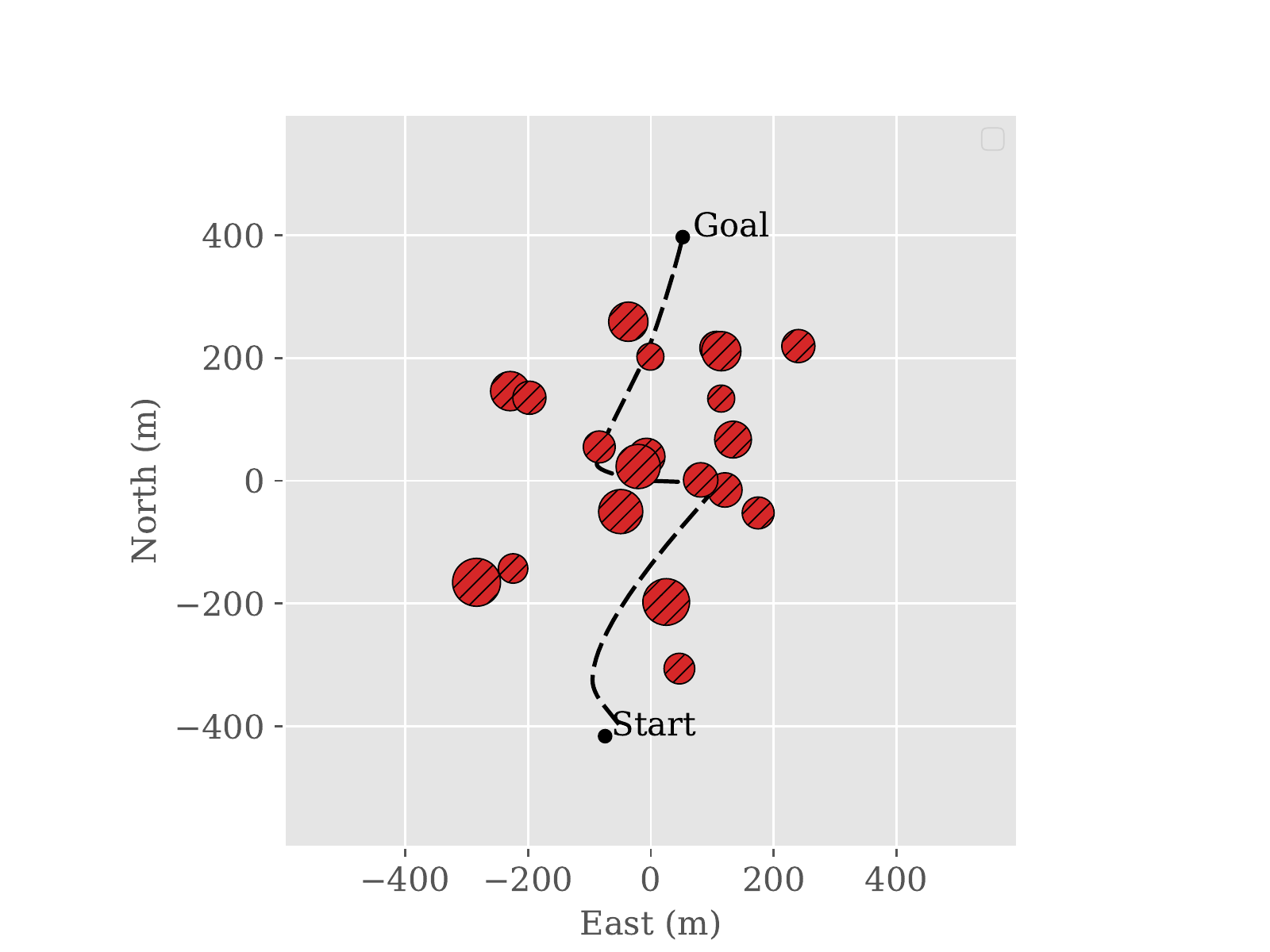}}
\end{subfigure}
\begin{subfigure}[]{0.49\linewidth}
    {\includegraphics[trim={2cm 0.7cm 3cm 1cm},clip,width=\linewidth]{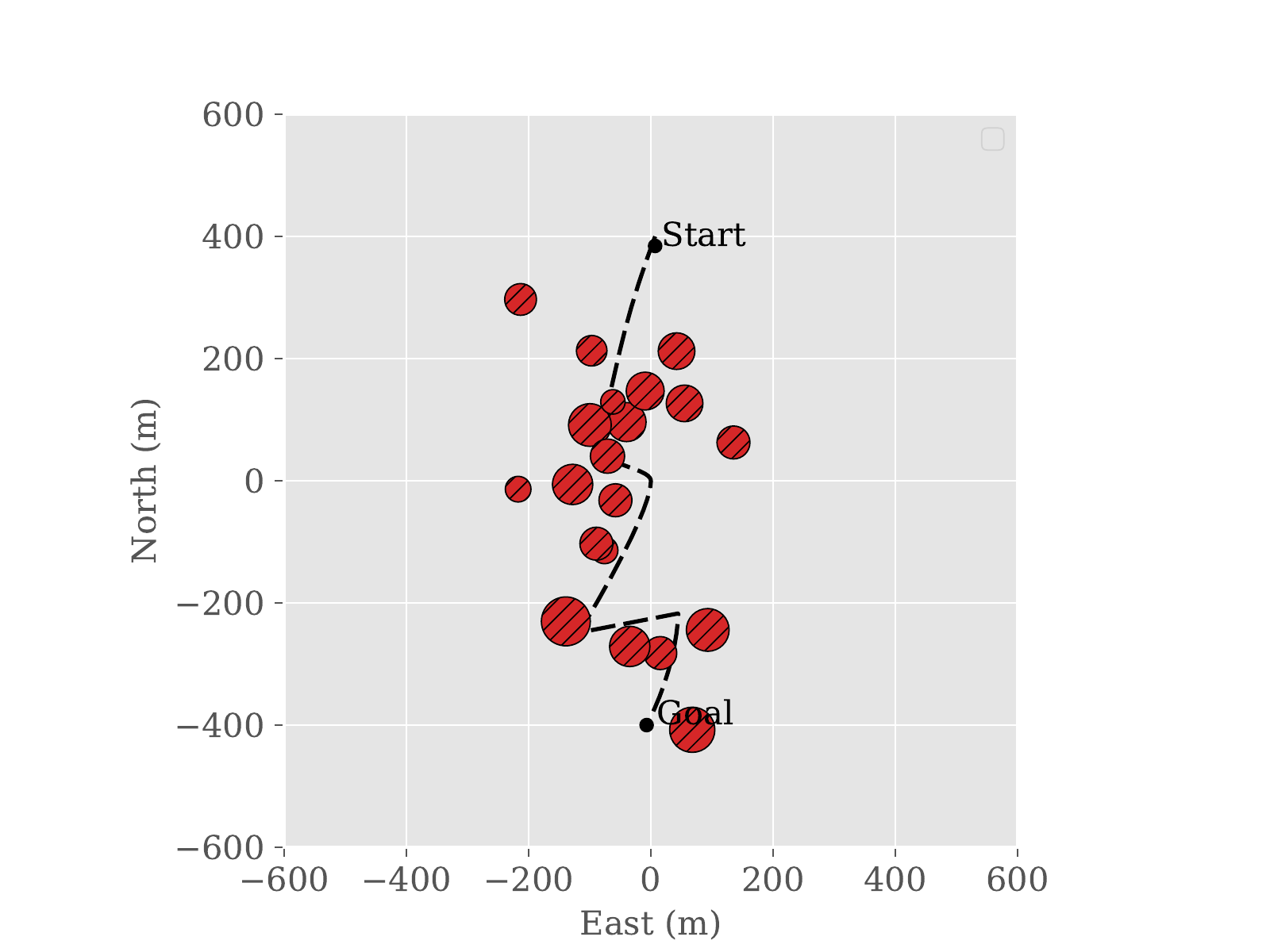}}
\end{subfigure}
\begin{subfigure}[]{0.49\linewidth}
    {\includegraphics[trim={2cm 0.7cm 3cm 1cm},clip,width=\linewidth]{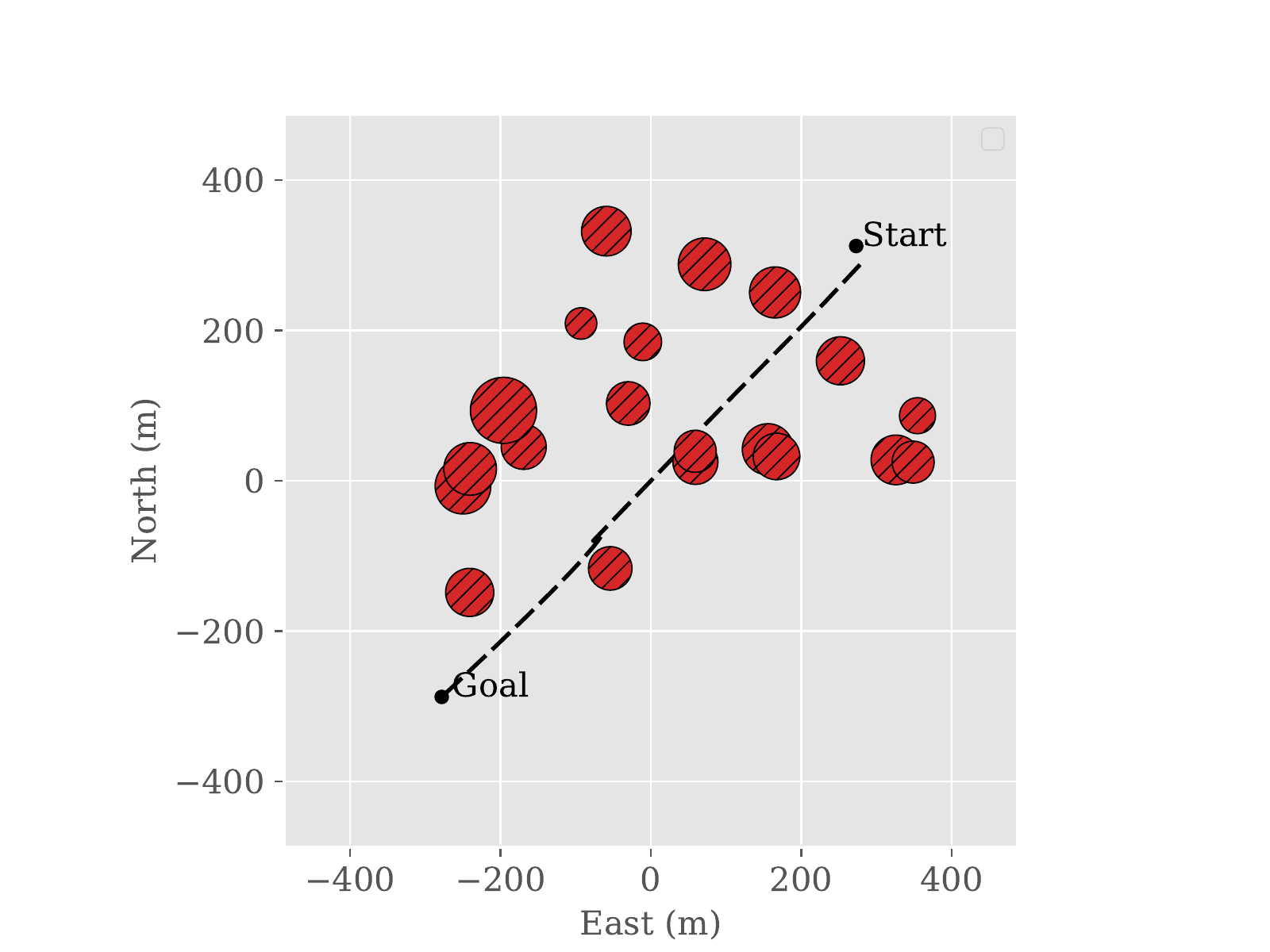}}
\end{subfigure}
\begin{subfigure}[]{0.49\linewidth}
    {\includegraphics[trim={2cm 0.7cm 3cm 1cm},clip,width=\linewidth]{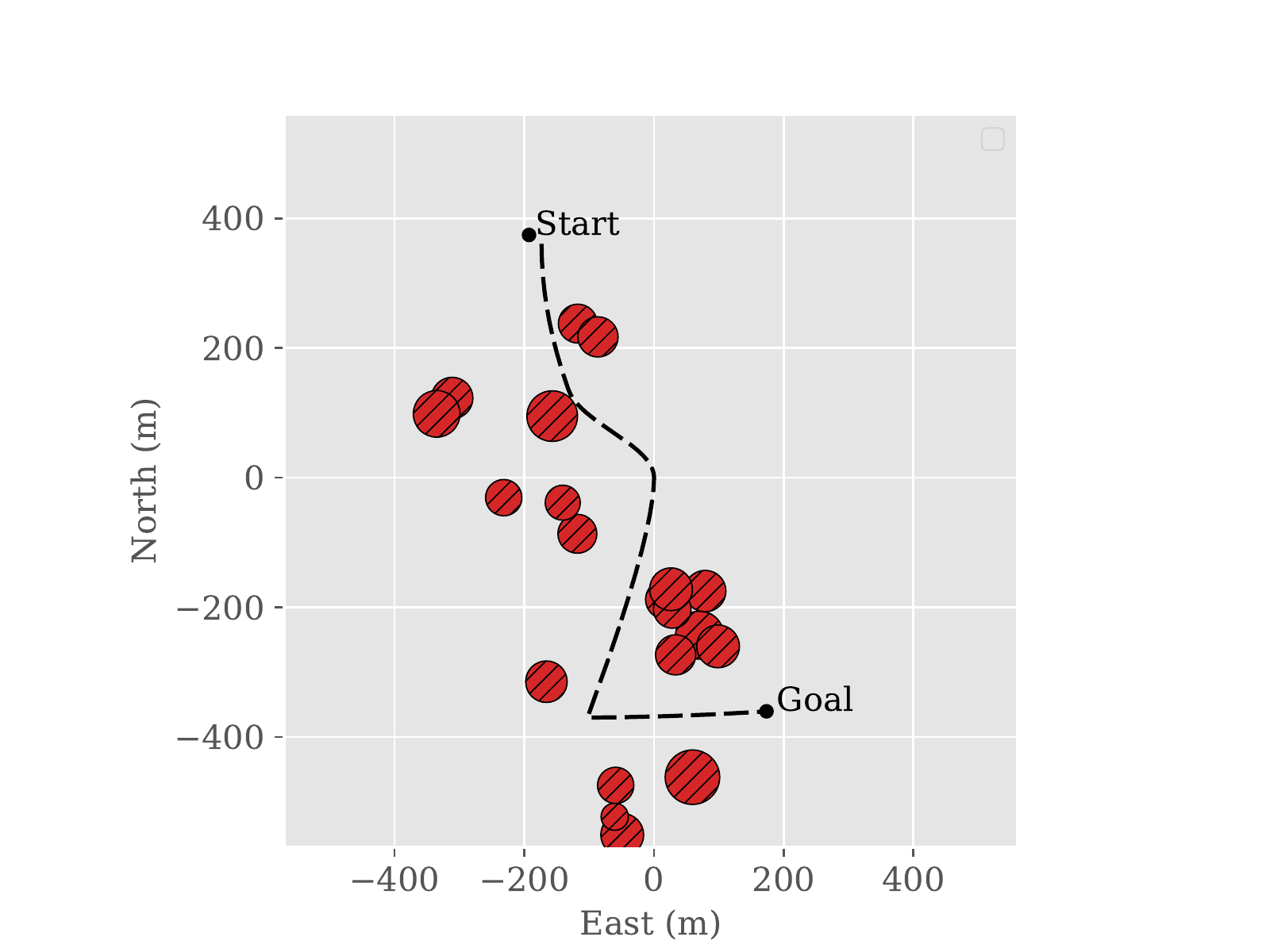}}
\end{subfigure}
\caption[Obstacle landscape samples]{Four random samples of the stochastically generated path following scenario. Note that the scenario difficulty is highly varying.}
\label{fig:landscape_samples}
\end{figure}

The environment in which we except the agent to perform is an ocean surface filled with obstacles, also containing an a priori known path that the agent is intended to follow while avoiding collisions. The vessel dynamics (ref. Section \ref{section:vessel_dynamics}) should, in fact, also be considered as a part of the environment, as it is outside of the agent's control. It is also critical that the environments in which the agent is trained pose a wide variety of challenges to the agent, so that the trained agent is able to generalize to unseen obstacle landscapes, potentially following a deployment on a vessel in the real world. Thus, we need a stochastic algorithm for generating training environments. If the environments are too easy or monotone (or a combination thereof), the agent will overfit to the training environments leading to undesired behavior when testing it in unseen, more complicated obstacle landscapes. For instance, if all obstacles are located very close to the path within the training environments, the trained agent may exhibit undesired behavior by always going around obstacles to avoid them, whereas an intelligent agent would simply ignore obstacles that are not in its way, in order to stay on track. Also, if the obstacle density is too low, it is unlikely that the agent would perform well in a high-obstacle-density environment. To this end we suggest the procedure outlined in Algorithm \ref{alg:scenario_creation} for generating new, independent training environments. Some randomly sampled environments generated from this algorithm can be seen in Figure \ref{fig:landscape_samples}. It is obvious that performing well within these environments (i.e. adhering to the planned path while avoiding collisions) necessitates a nontrivial guidance algorithm.

\begin{algorithm}[!htb]
    \scriptsize
    \begin{algorithmic}
    \Require
    \Statex Number of obstacles $N_o \in \mathbb{N}_0$
    \Statex Number of path waypoints $N_w \in \mathbb{N}_0$
    \Statex Path length $L_p \in \mathbb{N}_0$
    \Statex Mean obstacle radius $\mu_{r} \in \mathbb{R}^+$
    \Statex Obstacle displacement distance standard deviation $\sigma_{d} \in \mathbb{R}^+$
    \Procedure{GeneratePathColavEnvironment}{$N_o$, $N_w$, $L_p$, $\mu_{r}$, $\sigma_{d}$}
        \State Draw $\theta_{start}$ from $\textit{Uniform}(0, 2 \pi)$
        \State Path origin $\bm{p}_{start} \gets 0.5 L_p \left[ \cos{(\theta_{start})}, \sin{(\theta_{start})} \right]^T$
        \State Goal position $\bm{p}_{end} \gets -\bm{p}_{start}$
        \State Generate $N_w$ random waypoints between $\bm{p}_{start}$ and $\bm{p}_{end}$.
        \State Create smooth arc length parameterized path ${\bm{p}_p}(\bar{\omega}) = [x_p(\bar{\omega}), y_p(\bar{\omega})]^T$ using 1D Piecewise Cubic Hermite Interpolator (PCHIP) provided by Python library SciPy \cite{SCIPY}.
        \Repeat
        \State 
            Draw arclength $\bar{\omega}_{obst}$ from $\textit{Uniform}(0.1 L_p, 0.9 L_p)$.
        \State
            Draw obstacle displacement distance $d_{obst}$ from $\mathcal{N}(0, \sigma_d^2)$
        \State Path angle $\gamma_{obst} \gets \atantwo{({\bm{p}_p}'(\bar{\omega}_{obst})_2, {\bm{p}_p}'(\bar{\omega}_{obst})_1)}$ 
        \State Obstacle position $\bm{p}_{obst} \gets \bm{p}_p  (\bar{\omega}_{obst}) +  d_{obst} [ \cos{(\gamma_{obst} - \frac{\pi}{2})}, \sin{(\gamma_{obst} - \frac{\pi}{2})} ]^T $
        \State
            Draw obstacle radius $r_{obst}$ from $\textit{Poisson}(\mu_{r})$. 
        \State Add obstacle $(\bm{p}_{obst}, r_{obst})$ to environment
        \Until {$N_0$ obstacles are created}
    \EndProcedure
    \end{algorithmic}
\caption{Generate path with obstacles}
\label{alg:scenario_creation}
\end{algorithm}
In the current work the values of $N_o=20$, $N_w=\mathcal{U}(2,5)$, $L_p=400$, $\mu_r=30$, $\sigma_d=150$ (where $\mathcal{U}$ is the uniform distribition) were used.

\subsection{Agent}
Although the \textit{agent}, within the context of RL, can be considered to be the vessel itself, it is more accurate to look at it as the guidance mechanism controlling the vessel, as its operation is limited to outputting the control signals that steer the vessel's actuators. As discussed in Section \ref{section:vessel_model_3dof}, the available control signals are the propeller thrust $T_u$, driving the vessel forward, and the rudder angle $T_r$, inducing a yawing moment such that the vessel's heading changes. The RL agent's action, which it will output at each simulated time-step, is then defined as the vector $a = [T_u, T_r]^T$. Specifically, the action network, which we train by applying the PPO algorithm described in Section \ref{section:surrogate_objective}, will output the control signals following a forward pass of the current observation vector through the nodes of the neural network. Also, the value network is trained simultaneously, facilitating estimation of the state value function $V(s)$ which is used for GAE as described in Section \ref{section:gae}. Deciding what constitutes a state $s$ is of utmost importance; the information provided to the agent must be of sufficient fidelity for it to make rational guidance decisions, especially as the agent will be purely reactive, i.e. not be able to let previous observations influence the current action. At the same time, by including too many features in the state definition, we are at risk of overparameterization within the neural networks, which can lead to poor performance and excessive training time requirements \cite{goodfellowDL}. Thus, a compromise must be reached, ensuring a sufficiently low-dimensional observation vector while still providing a sufficiently rich observation of the current environment. Having separate observation features representing path following performance and obstacle closeness is a natural choice.

\subsubsection{Path following}
The agent needs to know how the vessel's current position and orientation aligns with the desired path. A few concepts often used for guidance purposes are useful in order to formalize this. First, we formally define the desired path as the one-dimensional manifold given by
\begin{equation}
\mathcal{P} \triangleq \left\{ \bm{p} \in \mathbb{R}^2 \;\mid\; \bm{p} = \bm{p}_p (\bar{\omega})\;\forall\; \bar{\omega} \in \mathbb{R}^+ \right\}
\end{equation}
\noindent Accordingly, for any given $\bar{\omega}$, we can define a local path reference frame $\{p\}$ centered at $\bm{p}_p (\bar{\omega})$ whose x-axis has been rotated by the angle
\begin{equation}
    \gamma_{p}(\bar{\omega}) \triangleq \atantwo{(y_{p}^{\prime}(\bar{\omega}), x_p^{\prime}(\bar{\omega}))}
\end{equation}
\noindent relative to the inertial NED-frame. Next, we consider the so-called look-ahead point $\bm{p}_p (\bar{\omega} + \Delta_{LA})$, where $\Delta_{LA} > 0$ is the look-ahead distance. In traditional path-following, look-ahead based steering, i.e. setting the look-ahead point direction as the desired course angle, is a commonly used guidance principle \cite{fossen11}. Based on the look-ahead point, we define the \textit{course} error, i.e. the course change needed for the vessel to navigate straight towards the look-ahead point, as
\begin{equation}
    \tilde{\chi}(t) \triangleq \atantwo{\left(\frac{y_p(\bar{\omega} + \Delta_{LA}) - y_p(\bar{\omega})}{x_p(\bar{\omega} + \Delta_{LA}) - x_p(\bar{\omega})}\right)} - \chi(t)
\end{equation}
\noindent where $\chi(t)$ is the vessel's current heading as defined in Section \ref{section:sname_state_variables}. Furthermore, given the current vessel position $\bm{p}(t)$ we can define the error vector $\bm{\epsilon}(t) \triangleq \left[s(t), e(t)\right]^T \in \mathbb{R}^2$, containing the \textit{along-track} error $s(t)$ and the \textit{cross-track} error $e(t)$ at time $t$, as
\begin{equation}
    \bm{\epsilon}(t) = \mathbf{R}_{z, -\gamma_p(\bar{\omega})} (\bm{p}(t) - \bm{p}_p(\bar{\omega}))
\end{equation}
\noindent \cite{BreivikFossen09}.
A natural approach for updating the path variable $\bar{\omega}$ is to repeatedly calculate the value that yields the closest distance between the path and the vessel using Newton's method. Here, the fact that Newton's method only guarantees a local optimum is a useful feature, as it prevents sudden path variable jumps given that the previous path variable value is used as the initial guess \cite{martinsen2018}. Another approach is to update the path variable according to the differential equation
\begin{equation}
    \dot{\bar{\omega}} = \sqrt{u^2 + v^2} \cos{\tilde{\chi}(t)} - \gamma_{\hat{\omega}} s(t)
\end{equation}
\noindent where the along-track error coefficient $\gamma_{\hat{\omega}} > 0$ ensures that the absolute along-track error $|s(t)|$ will decrease. As this method is computationally faster, we choose to use it in our Python implementation. More specifically in the current work $\gamma_{\hat{\omega}} =0.05$ and $\Delta_{LA}=100m$. 


\begin{figure}[pos=!ht]
\centering
\begin{subfigure}{0.8\linewidth}
    \includegraphics[width=\linewidth]{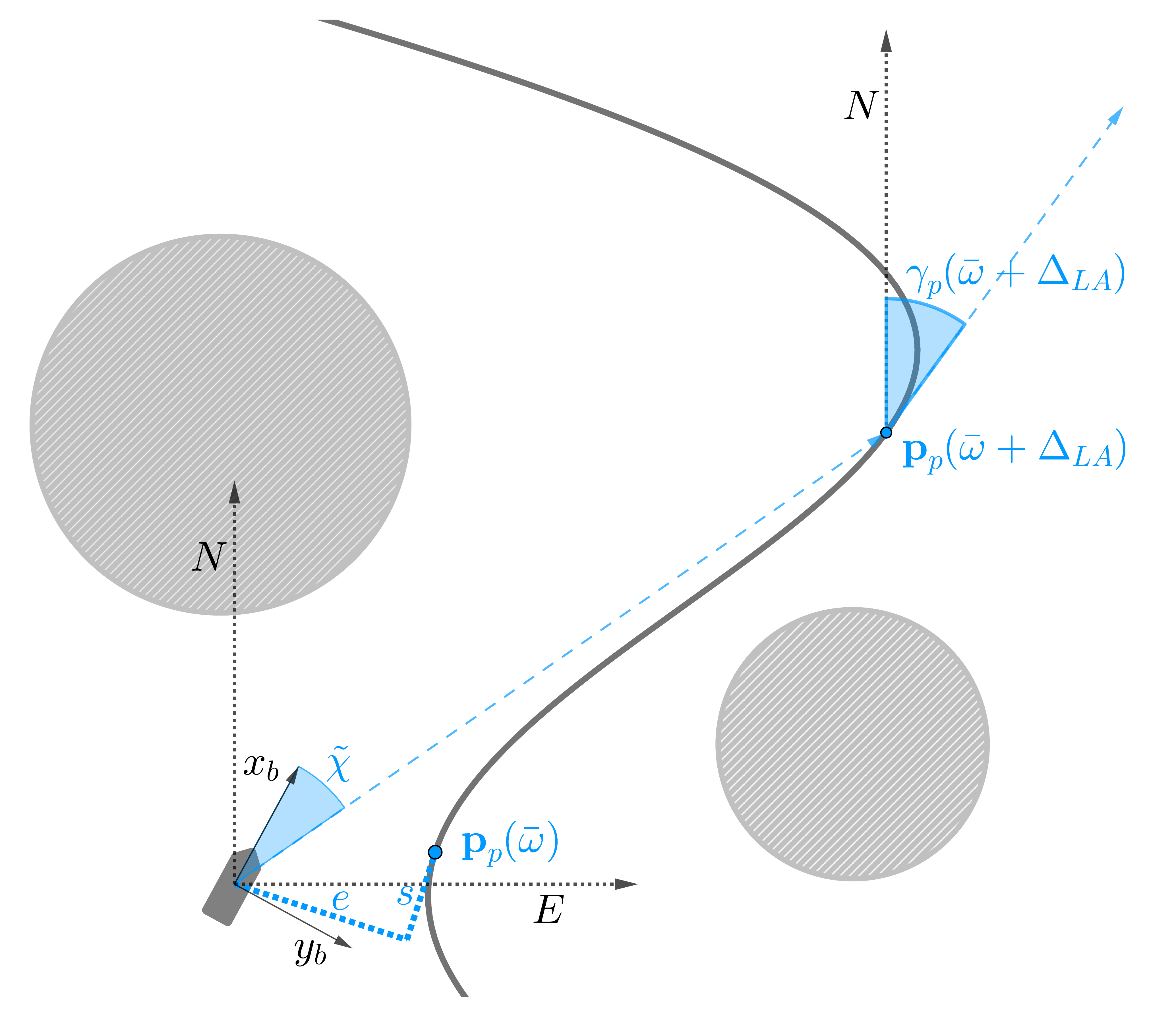}
    \subcaption{Distances and angles for path following}
    \label{vessel_path_following}
\end{subfigure}
\begin{subfigure}{0.8\linewidth}
    \includegraphics[width=\linewidth]{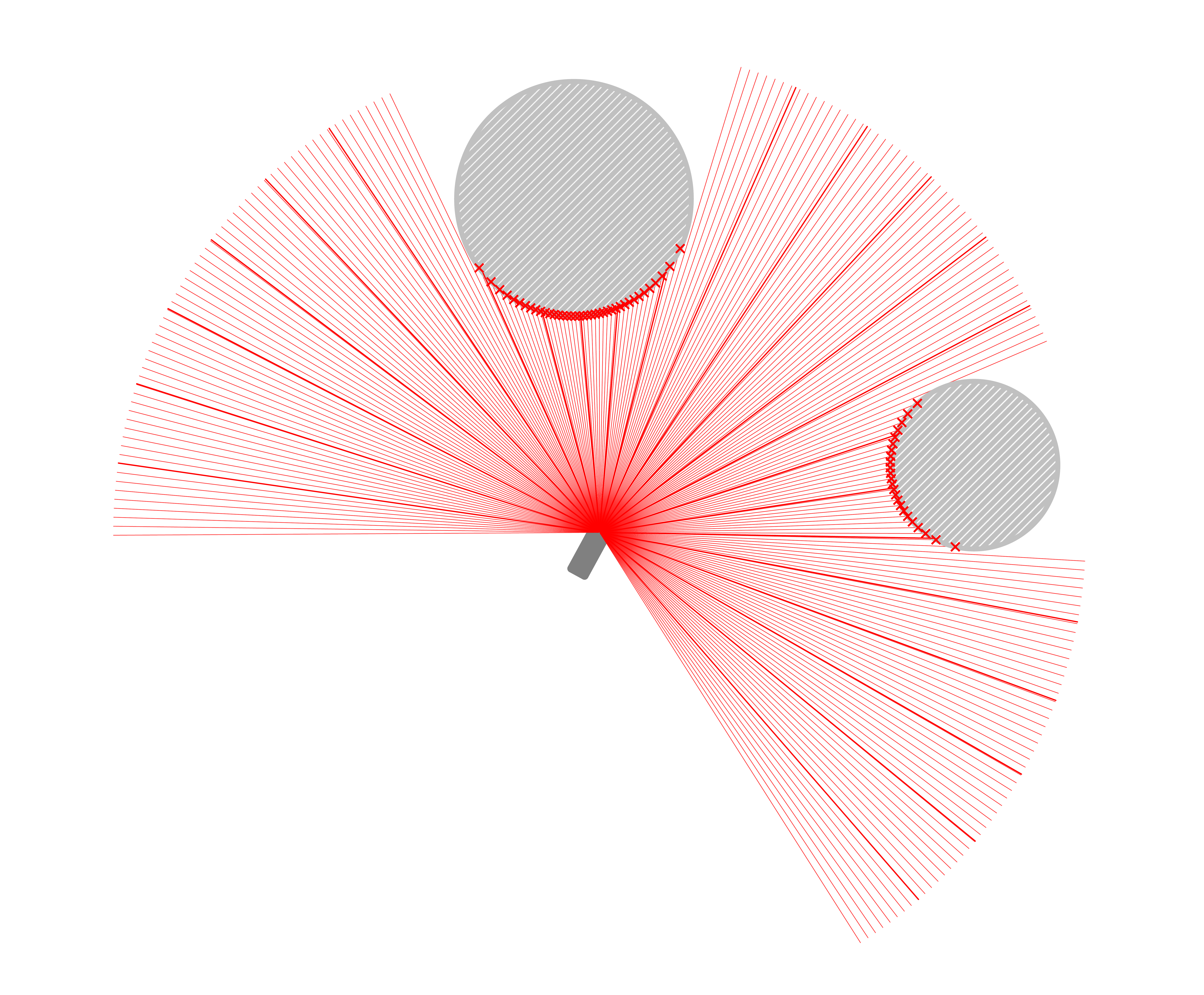}
    \subcaption{$N = 225$ rangefinder sensors partitioned into $d = 25$ sectors}
    \label{segmented_lidars}
\end{subfigure}
\caption{Illustrations showing the parameters for path following and collision avoidance. (a) shows the cross-track error $e$, along-track error $s$, heading error $\tilde{\chi}$, path reference point $\bm{p}_p(\bar{\omega})$, look-ahead point $\bm{p}_p(\bar{\omega} + \Delta_{LA})$ and look-ahead path tangential angle $\gamma_p (\bar{\omega} + \Delta_{LA})$. In (b), the sensors are arranged in sectors, where the sensor measurements are pooled into a scalar values.}
\end{figure}


\subsubsection{Obstacle detection}
\label{section:sensormeasurements}
Using rangefinder sensors as the basis for obstacle avoidance is a natural choice, as a reactive navigation system applied to a real-world vessel typically would entail either such a solution or a camera-based one. This realistic approach should enable a relatively straightforward transition from the simulated environment to a real one, given the availability of common rangefinder sensors such as lidar, radar or sonar.

In the setup used, $N = 225$ sensors with a total visual span of $S_s = \frac{4 \pi}{3}$ radians ($240$ degrees) are arranged in the trivial manner illustrated in Figure \ref{segmented_lidars}. The sensors are assumed to have a range of $S_r = 150$ meters, which was deemed sufficient given the relatively small size of the vessel. Obviously, with regards to the number of sensors, one must consider the trade-off between computation speed and sensor resolution. In the experiments conducted in this research project, $225$ sensors were chosen, even if it is likely that a much lower number of sensors would yield similar performance. With regards to the visual span, it could be argued that providing $180$ degree vision would be sufficient to achieve satisfactory collision avoidance, given the precondition of static obstacles. However, in the interest of avoiding sub-optimal performance due to a restrictive sensor suite configuration, the conservative choice of having $24$0 degree vision was made. 

Even if, in theory, a sufficiently large neural network is capable of representing any function with any degree of accuracy, including satisfactory mappings from sensor readings to collision-avoiding steering maneuvers in our case, there are no guarantees for neither the feasibility of the required network size nor the convergence of the optimization algorithm used for training to the optimal network weights \cite{goodfellowDL}. Thus, forcing the action network to output the control signal based on $225$ sensor readings (as well as the features intended for path-following) is unlikely to be a viable approach, given the complexity required for any satisfactory mapping between the full sensor suite to the steering signal. Instead, we propose three approaches for transforming the sensor readings into a reduced observation space from which a satisfactory policy mapping should be easier to achieve. As illustrated in Figure \ref{segmented_lidars}, this involves partitioning the sensor suite into $d$ disjoint sensor sets, hereafter referred to as \textit{sectors}. First, we define the sensor \textit{density} $n$ as the number of sensors contained by one sector: $n \triangleq \frac{N}{d}$


Each sector is made up of neighboring sensors, so we can formally define the $k^{th}$ sector, which we denote by $\mathcal{S}_k$, as 
\begin{equation}
    \mathcal{S}_k \triangleq \left\{ x_{(k - 1) n + 1}, \dots, x_{k n} \right\}
\end{equation}
where $x_i$ refers to the $i^{th}$ sensor measurement according to a counter-clockwise indexing direction. This partitioning, which assumes that $N$ is a multiple of $d$, is illustrated in Figure \ref{segmented_lidars}.

\begin{figure}[pos=!ht]
\centering
\begin{subfigure}{0.49\linewidth}
    {\includegraphics[width=\linewidth, trim={8cm 8cm 6.5cm 7cm}, clip]{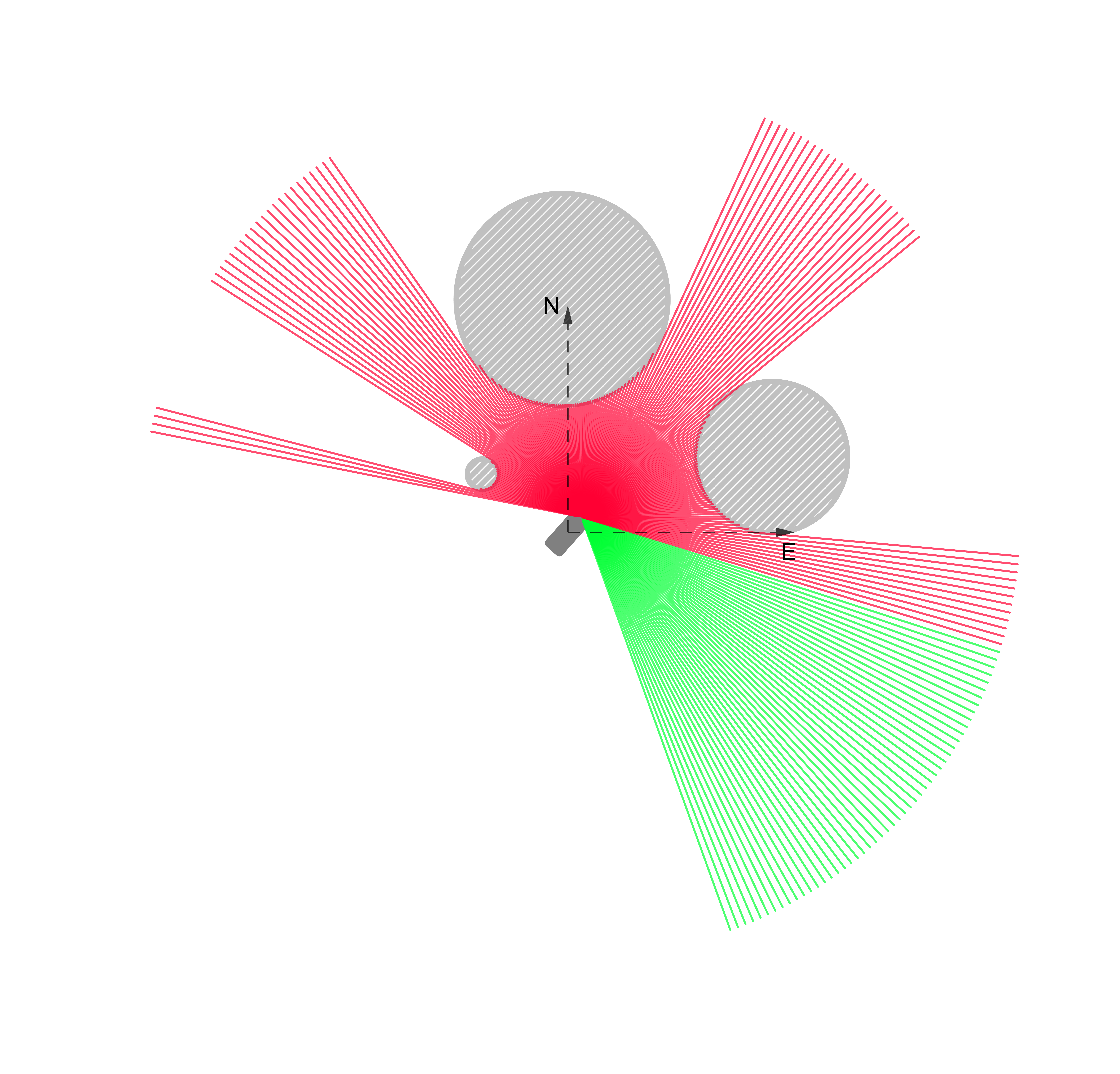}}
    \subcaption{Min pooling}
    \label{fig:minpooling}
\end{subfigure}
\begin{subfigure}{0.49\linewidth}
    {\includegraphics[width=\linewidth, trim={8cm 8cm 6.5cm 7cm}, clip]{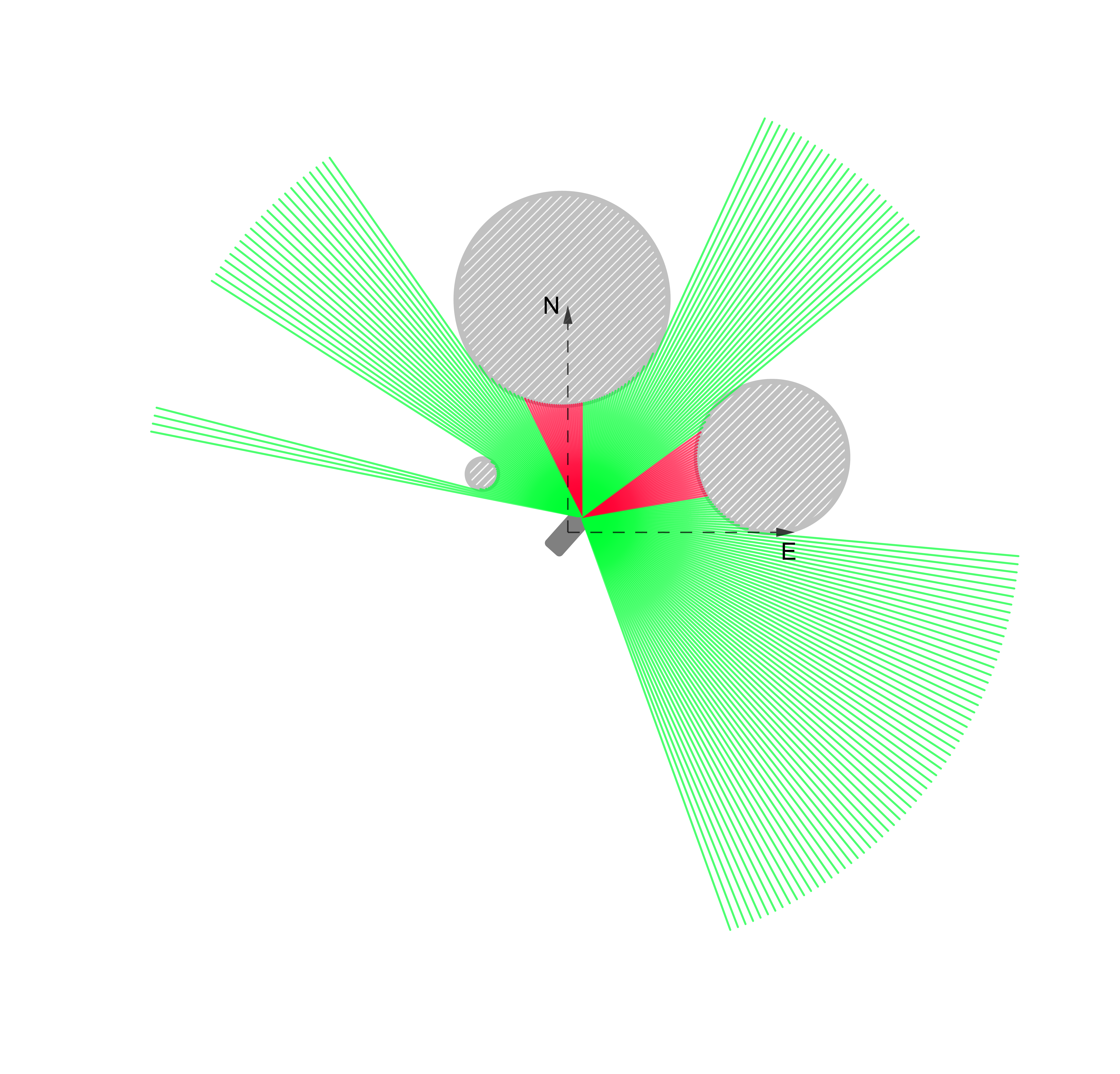}}
    \subcaption{Max pooling}
    \label{fig:maxpooling}
\end{subfigure}
\begin{subfigure}{0.49\linewidth}
    {\includegraphics[width=\linewidth, trim={8cm 8cm 6.5cm 7cm}, clip]{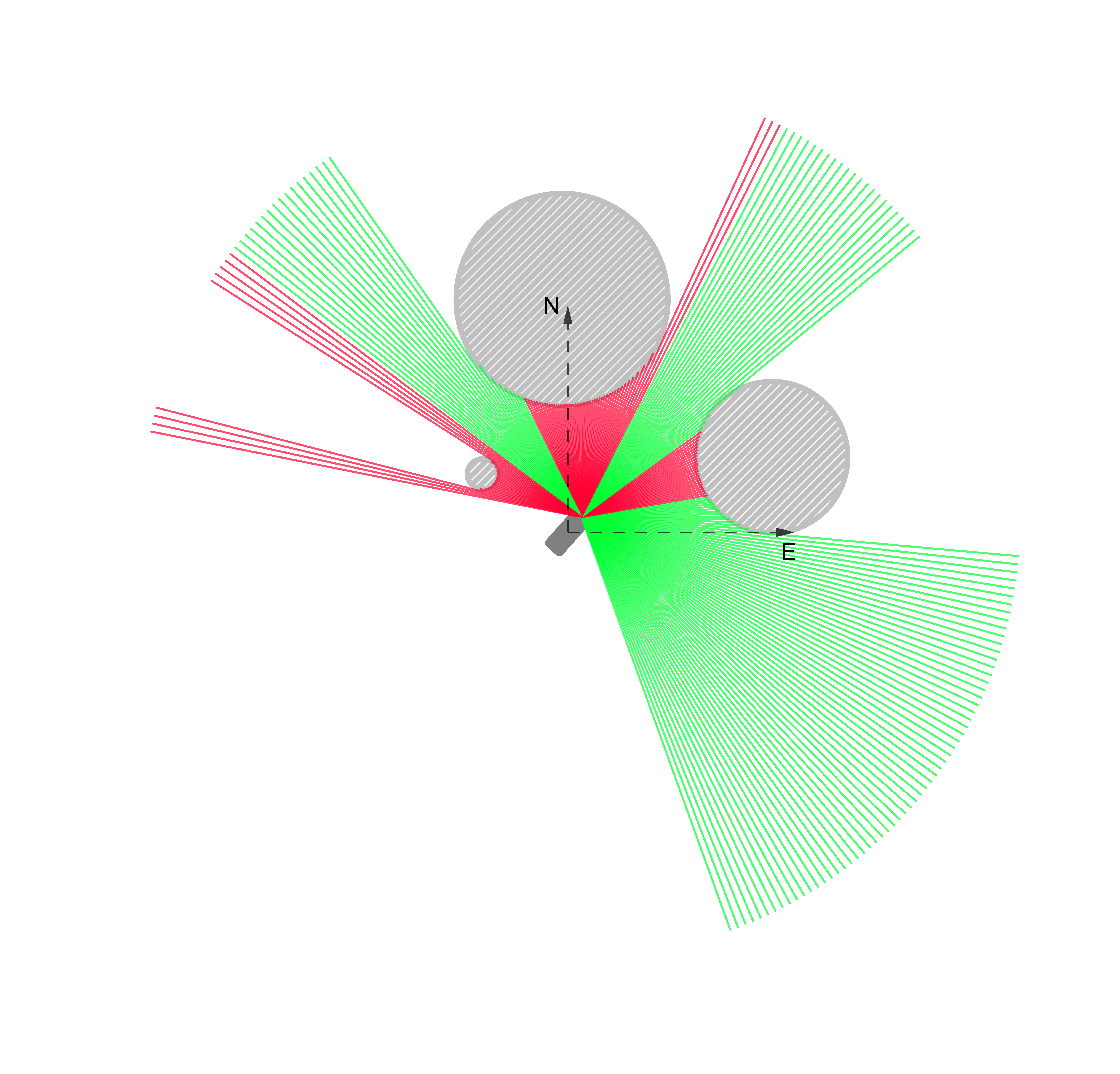}}
    \subcaption{Feasibility pooling}
\end{subfigure}
\caption{Pooling techniques for sensor dimensionality reduction. For the sectors colored green, the maximum distance $S_r$ was outputted. It is obvious that min-pooling yields an overly restrictive observation vector, effectively telling the agent that a majority of the travel directions are blocked. On the other hand, max pooling yields overly optimistic estimates, potentially leading to dangerous situations.}
\end{figure}

Based on the concept of partitioning the sensor suites into sectors, we then seek to reduce the dimensionality of our observation vector. Instead of including each individual sensor measurement $x_i$ in it, we provide a single scalar feature for each sector $\mathcal{S}_k$, effectively summarizing the local sensor readings within the sector. The resulting dimensionality reduction is quite significant; instead of having $N$ sensor measurements in the observation vector, we now have only $d$ features. What remains is the exact computation procedure by which a single scalar is outputted based on the current sensor readings within each sector. Always returning the minimum sensor reading within the sector, in the following referred to as \textit{min pooling}, i.e. outputting the shortest measured obstacle distance within the sector, is a natural approach which yields a conservative and thereby safe observation vector. As can be seen in Figure \ref{fig:minpooling}, however, this approach might be overly restrictive in certain obstacle scenarios, where feasible passings in between obstacles are inappropriately overlooked. However, even if the opposite approach (\textit{max pooling}) solves this problem, it is straight-forward to see, e.g. in Figure \ref{fig:maxpooling} by considering the fact that the presence of a small, nearby obstacle in the leftmost sector is ignored, that it might lead to dangerous navigation strategies.

To alleviate the problems associated with min and max pooling mentioned above a new approach is required. A natural approach is to compute the maximum feasible travel distance within the sector, taking into account the location of the obstacle sensor readings as well as the width of the vessel. This requires us to iterate over the sensor reading in ascending order corresponding to the distance measurements, and for each resulting distance level check whether it is feasible for the vessel to advance beyond this level. As soon as the widest opening available within a distance level is deemed too narrow given the width of the vessel, the maximum feasible distance has been reached. A pseudocode implementation of this algorithm is provided as Algorithm \ref{alg:feasibility_pooling}.


\begin{algorithm}
\scriptsize
    \begin{algorithmic}
    \Require
    \Statex Vessel width $W \in \mathbb{R}^+$
    \Statex Total number of sensors $N \in \mathbb{N}$
    \Statex Total sensor span $S_s \in [0, 2 \pi]$ 
    \Statex Sensor rangefinder measurements for current sector $\bm{x} = \{x_1, \dots, x_n\}$
    \Procedure{FeasibilityPooling}{$\bm{x}$}
        \State Angle between neighboring sensors $\theta \gets \frac{S_s}{N-1}$
        \State Initialize $\mathcal{I}$ to be the indices of $\bm{x}$ sorted in ascending order according to the measurements $x_i$
        \For{$i \in \mathcal{I}$}
            \State Arc-length $d_i \gets \theta x_i$
            \State Opening-width $y \gets d_i/2$
            \State Opening was found $s_i \gets$ $false$
            \For{$j \gets 0$ to $n$}
                \If{$x_j > x_i$}
                    \State $y \gets y + d_i$
                    \If{$y > W$}
                        \State $s_i \gets$ $true$
                        \State \textbf{break}
                    \EndIf
                \Else
                    \State $y \gets y + d_i/2$
                    \If{$y > W$}
                        \State $s_i \gets$ $true$
                        \State \textbf{break}
                    \EndIf
                    \State $y \gets 0$
                \EndIf
            \EndFor
            \If{$s_i$ is $false$}
                \Return $x_i$
            \EndIf
        \EndFor
    \EndProcedure
    \end{algorithmic}
\caption[Feasibility pooling algorithm]{Feasibility pooling for rangefinder}
\label{alg:feasibility_pooling}
\end{algorithm}

\begin{figure}[pos=!ht]
\centering
\begin{subfigure}{0.8\linewidth}
    \includegraphics[width=\linewidth]{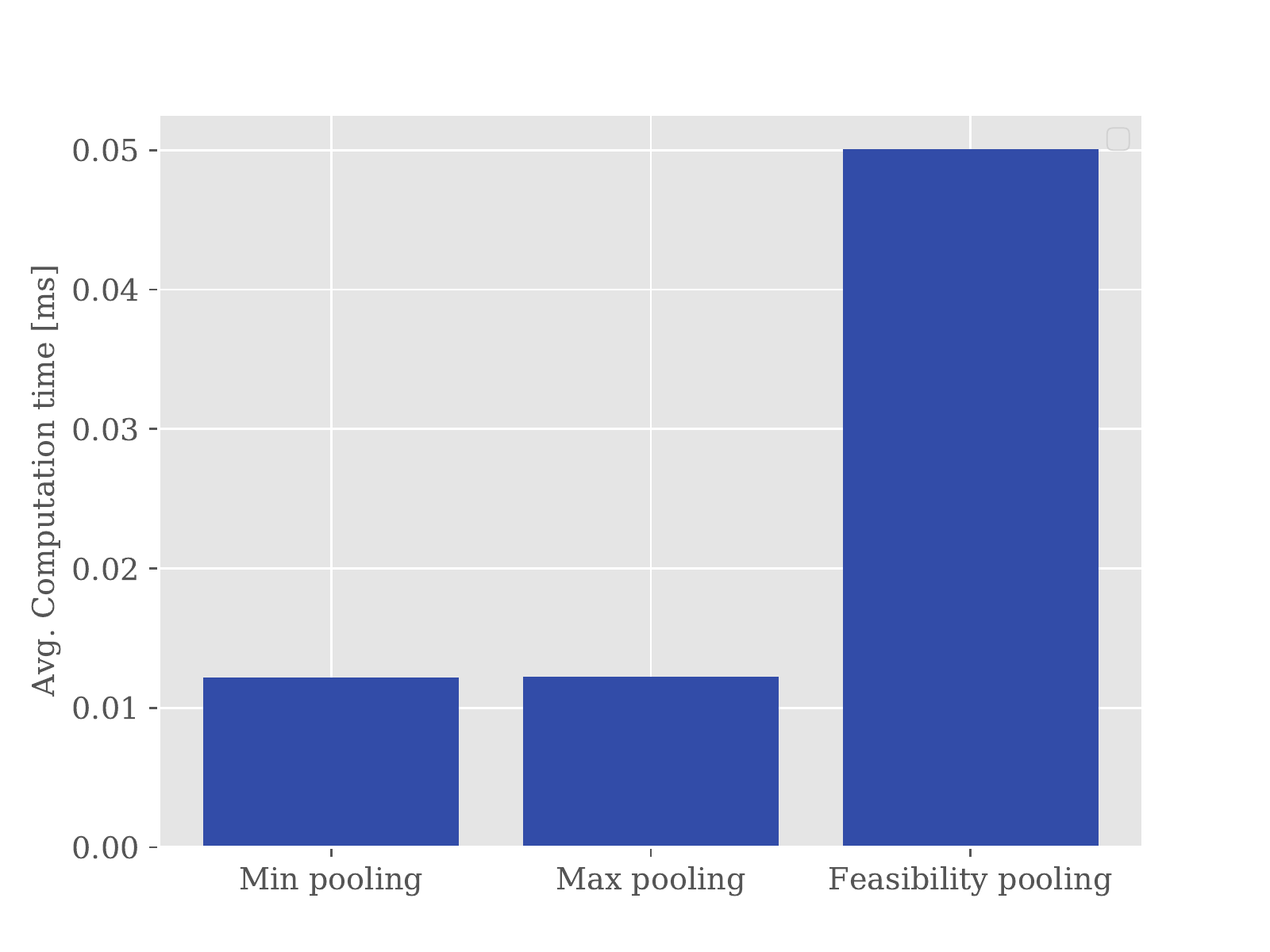}
    \subcaption{Average per-sector computation time for pooling methods when $n=9$ }
    \label{fig:pooling_computation_time}
\end{subfigure}
\begin{subfigure}{0.8\linewidth}
    \includegraphics[width=\linewidth]{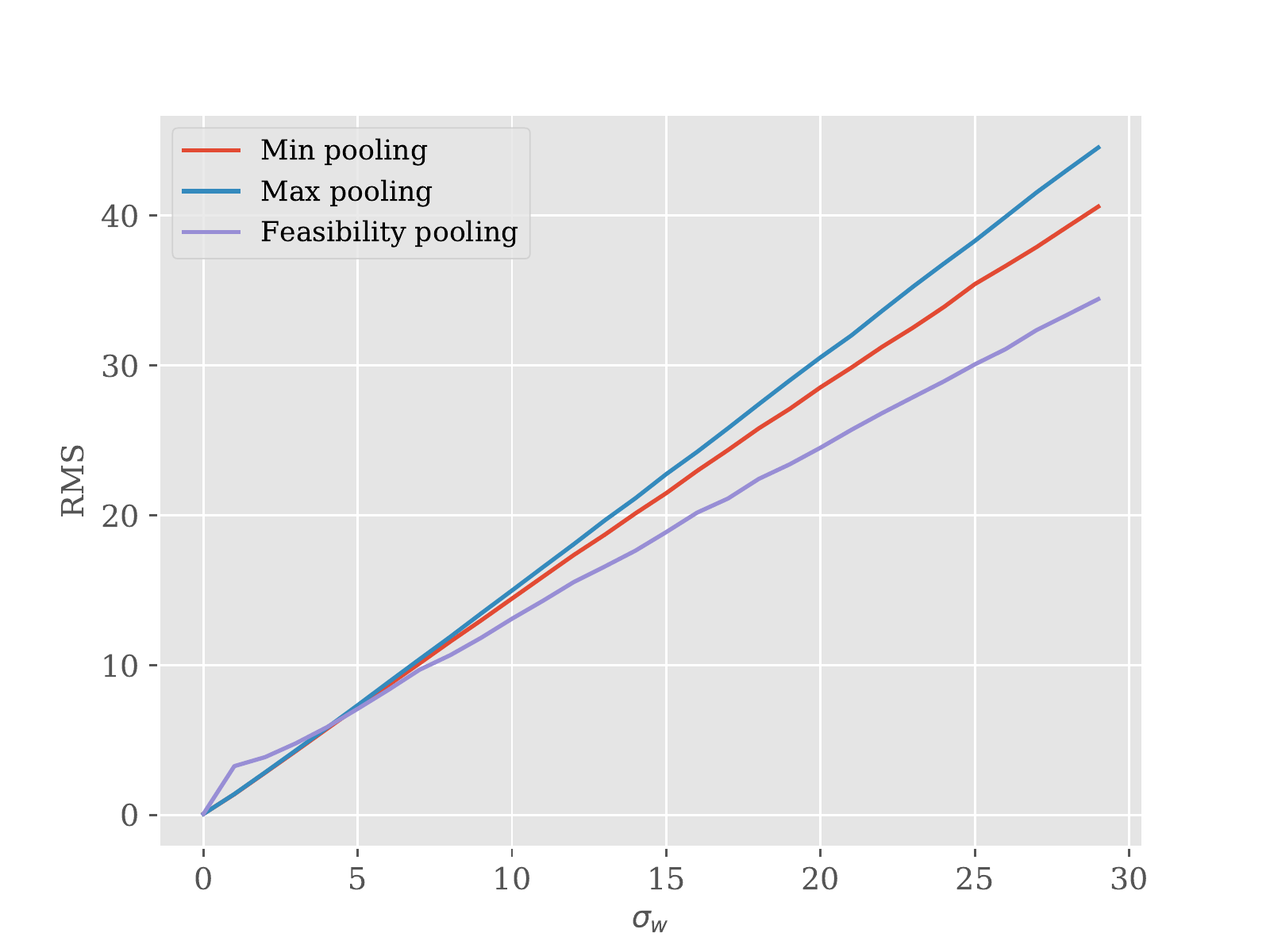}
    \subcaption{Robustness metric for pooling methods for $\sigma_w \in \{1, \dots, 30\}$ }
    \label{fig:pooling_robustness}
\end{subfigure}
\caption[Sensor pooling methods]{Computational time and robustness of the different pooling approaches. The noise-affected measurements were clipped at zero to avoid negative values.}
\end{figure}

Having a runtime complexity of $\mathcal{O}(d n^2)$ when executed on the entire sensor suite, the feasibility pooling approach is slower than simple max or min pooling, both having the runtime complexity $\mathcal{O}(d n)$. However, in the simulated environment, the increased computation time, which is reported through empirical estimates in Figure \ref{fig:pooling_computation_time} for $n = 9$, is negligible compared to the time needed to compute the interception points between the rangefinder rays and the obstacles. 

Another interesting aspect to consider when comparing the pooling methods, is the sensitivity to sensor noise. A compelling metric for this is the degree to which the pooling output differs from the original noise-free output when normally distributed noise with standard deviation $\sigma_w$ is applied to the sensors. Specifically, we report the root mean square of the differences between the original pooling outputs and the outputs obtained from the noise-affected measurements. The results for $\sigma_w \in \{1, \dots, 30\}$ are presented in Figure \ref{fig:pooling_robustness}. Evidently, the proposed feasibility method for pooling is slightly more robust than the other variants.


\begin{table}[pos=h]
	\begin{tabular}{lll}
		\hline
		Parameter & Description & Value \\
		\hline
		$U_{max}$ & Maximum vessel speed & $2$ m/s \\
		$W$ & Vessel width & $4$ m \\
		$N$ & Number of sensors & $225$ \\
		$S_s$ & Total visual span of sensors & $240^{\circ}$\\
		$S_r$ & Maximum rangefinder distance & $150$ m\\
		$d$ & Number of sensor sectors & $25$ \\\hline
	\end{tabular}
	\caption{Sensor configuration}
	\label{tab:vesselconfig_obstdect}
\end{table}

\subsection{Rewards}
Any RL agent is motivated by the pursuit of maximum reward.  Thus, designing the reward function $r(t)$ is paramount to the agent exhibiting the desired behavior after training. Given the dual nature of our objective, which is to follow the path while avoiding obstacles along the way, it is natural to reward the agent separately for its performance in these two domains. Thus, we introduce the reward terms $r_{pf}(t)$ and $r_{oa}(t)$, being the reward components at time $t$ representing the path-following and the obstacle-avoiding performance, respectively. Also, we introduce the weighting coefficient $\lambda \in [0, 1]$ to regulate the trade-off between the two competing objectives, leading to the preliminary reward function 
\begin{equation}
r(t) =  \lambda r_{pf}(t) + (1 - \lambda) r_{oa}(t)
\end{equation}

\subsubsection{Path following performance}
\begin{figure}[pos=!ht]
\centering
\begin{subfigure}{0.8\linewidth}
    \centering
    \includegraphics[width=\textwidth]{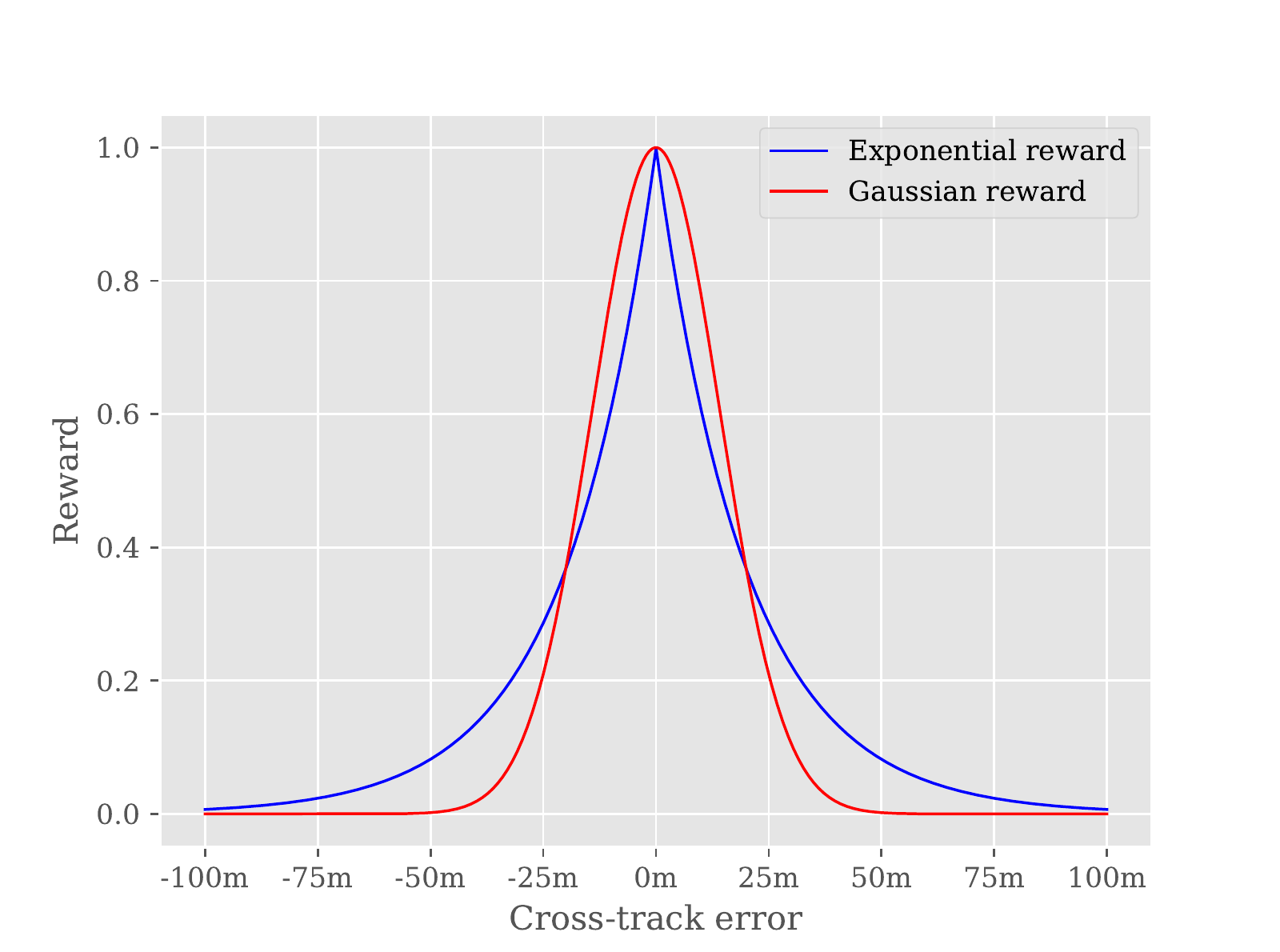}
    \subcaption{Cross-section of the path-following reward landscape assuming path-tangential full-speed motion}
    \label{fig:reward_landscape_2d}
\end{subfigure}
\begin{subfigure}{0.9\linewidth}
    \centering
    \includegraphics[width=\textwidth]{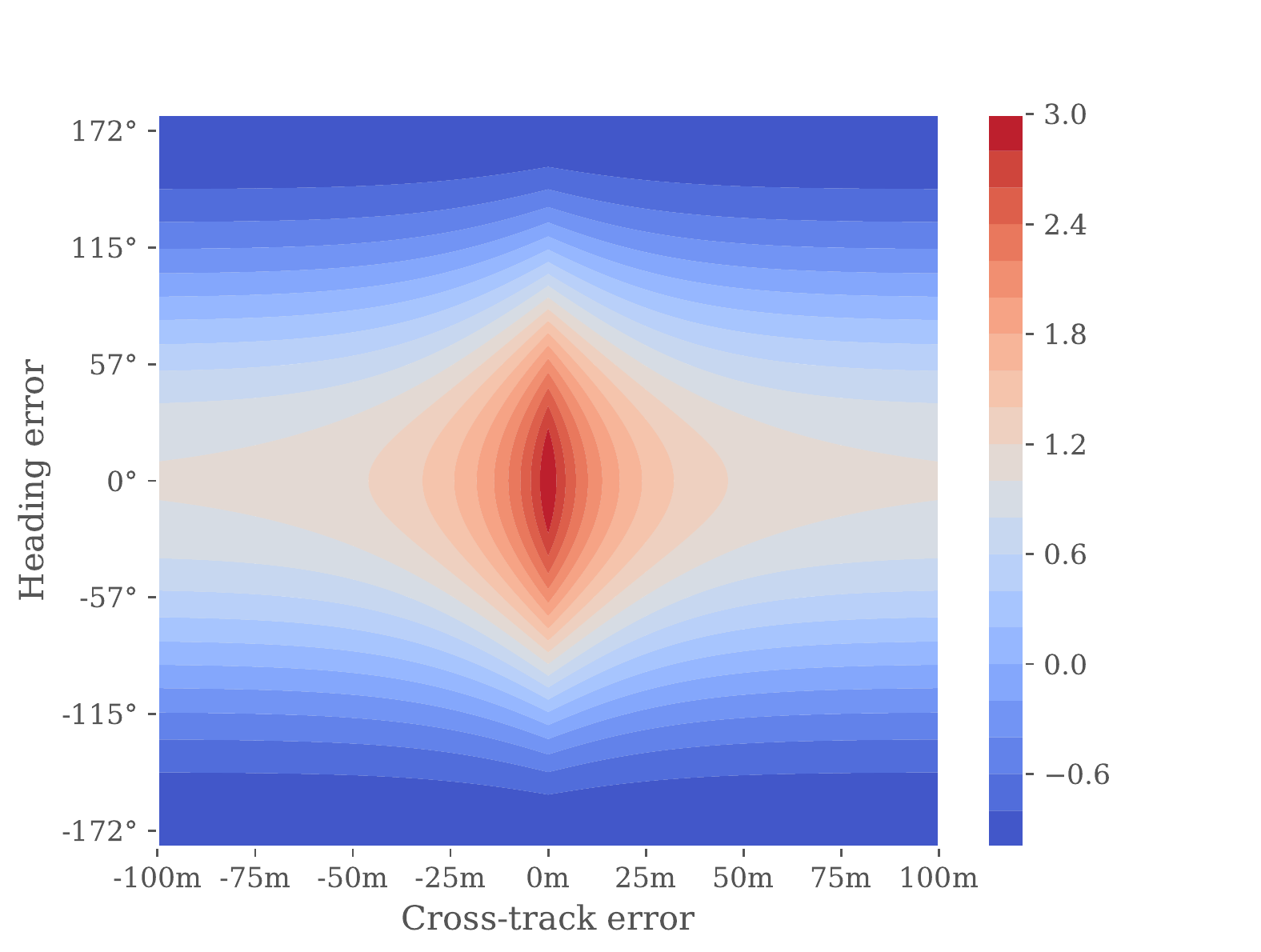}
    \subcaption{Path-following reward function assuming full-speed motion}
    \label{fig:reward_landscape_3d}
\end{subfigure}
\caption{Cross-section and level curves for the path-following reward function for $\gamma_{e} = 0.05$}
\end{figure}

A reasonable approach to incentivize adherence to the desired path is to reward the agent for minimizing the absolute cross-track error $e(t)$. In \cite{martinsen2018}, a Gaussian reward function centered at $e(t)=0$ with some reasonable standard deviation $\sigma_e$ is used for this purpose. However, based on Figure \ref{fig:reward_landscape_2d}, we argue that the exponential $e^{-\gamma_{e} |y_{e}(t)|}$
has slightly more reasonable characteristics for this purpose due to its fatter tails, thus rewarding the agent for a slight improvement to an unsatisfactory location.  


However, this alone does not reflect our desire for the agent to actually make progress along the path. This can be achieved by multiplying by the velocity component in the desired course direction given by $\sqrt{u^2 + v^2} \cos{\Tilde{\chi}(t)}$, effectively yielding negative rewards if the agent is tracking backwards, and zero reward if it is vessel course in a direction perpendicular to the path. Finally, we note that, if the agent is standing still, or if the course error is $\pm 90^{\circ}$, it will receive zero reward regardless of the cross-track error, which is undesired. Similarly, when the cross-track error grows large, it receive zero reward regardless of the speed or course error. Thus, we add constant multiplier terms $1$ and end up with the path-following reward function
\begin{equation}
r_{pf}(t) = -1 +\left(\tfrac{\sqrt{u^2 + v^2}}{U_{max}} \cos{\Tilde{\chi}(t)} + 1 \right) \left(e^{-\gamma_{e} |y_{e}(t)|} + 1\right) 
\end{equation}
\noindent where $U_{max}$ is the maximum vessel speed.


\begin{remark}
Note that, for added flexibility, it is possible to replace the $1$ multipliers by some customizable coefficients. However, for the sake of parametric simplicity, we decide to use $1$.
\end{remark}

\subsubsection{Obstacle avoidance performance}

\begin{figure}[pos=!htb]
  \centering
    \includegraphics[width=\linewidth]{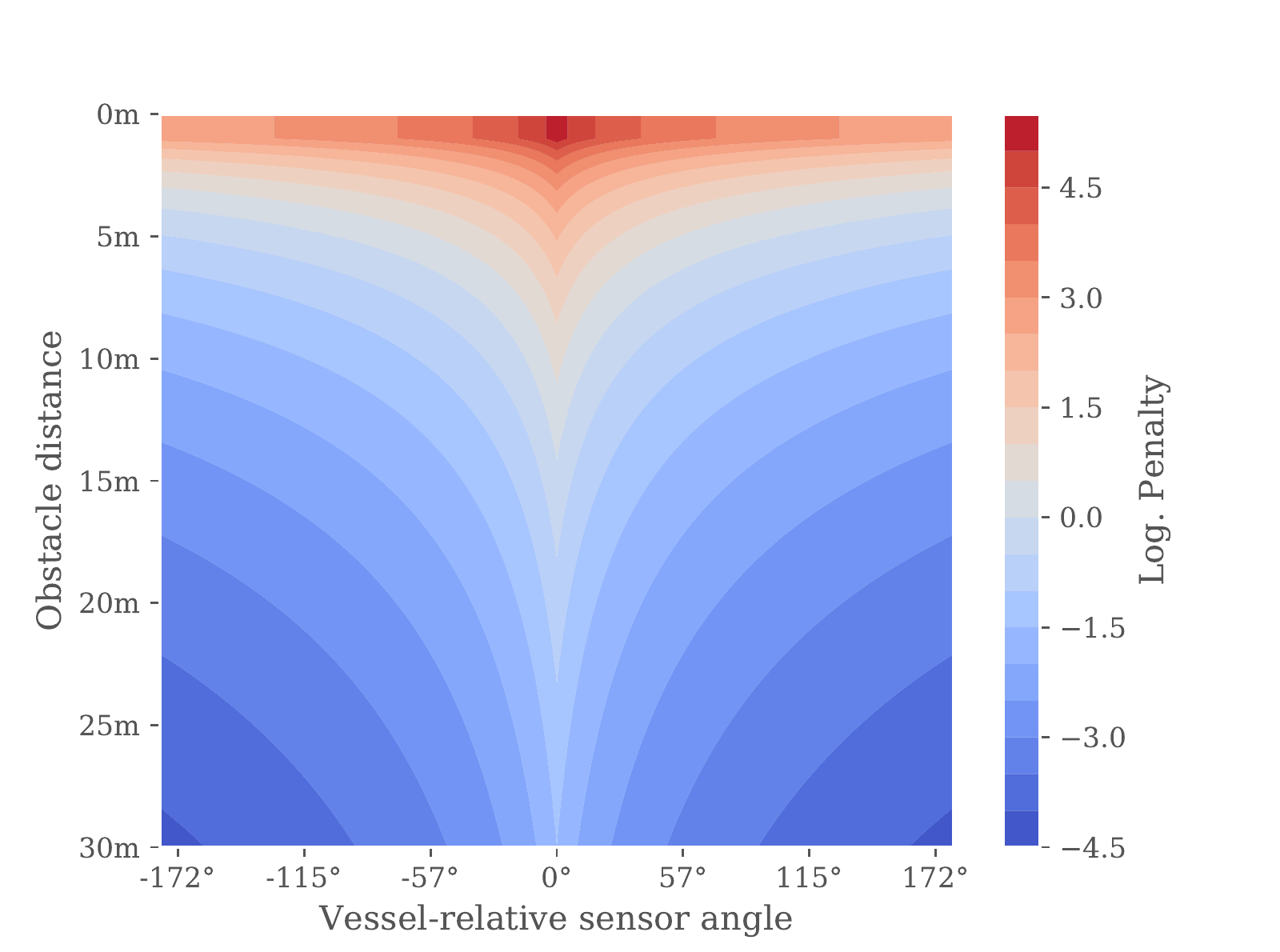}
 \caption[Obstacle closeness penalty landscape]{Obstacle closeness penalty as a function of vessel-relative sensor angle and obstacle distance, imposing a maximum penalty for obstacles located right in front of the vessel.}
\label{fig:obst_reward_landscape}
\end{figure}

In order to encourage obstacle-avoiding behavior, penalizing the agent for the \textit{closeness} of nearby obstacles in a strictly increasing manner seems natural. Having access to the sensor measurements outlined in Section \ref{section:sensormeasurements} at each timestep, we use these as surrogates for obstacle distances through which the agent is penalized. By noting that the severity of obstacle closeness intuitively does not increase linearly with distance, but instead increases in some more or less exponential manner, and that the severity of obstacle closeness depends on the orientation of the vessel with regards to the obstacle in such a manner that obstacles located behind the vessel are of much lower importance than obstacles that are right in front of the vessel, is it easy to see that the term ${\left({1 + |\gamma_{\theta}\theta_i|}\right)}^{-1} {(\gamma_x {\max{(x_i, \epsilon_x)}}^2)}^{-1}$, where $\theta_i$ is the vessel-relative angle of sensor $i$ such that a forward-pointing sensor has angle $0$, exhibits the desirable properties for penalizing the vessel based on the $i^{th}$ sensor reading. This reward function is plotted in Figure \ref{fig:obst_reward_landscape}.

In order to to cancel the dependency on the specific sensor suite configuration, i.e. the number of sensors and their vessel-relative angles, that arises when this penalty term is summed over all sensors, we use a weighted average to define our obstacle-avoidance reward function such that  
\begin{equation}
r_{oa}(t) = -\frac
{\textstyle \sum_{i=1}^{N}{{\left({1 + |\gamma_{\theta}\theta_i|}\right)}^{-1} {(\gamma_x {\max{(x_i, \epsilon_x)}}^2)}^{-1} }}
{\textstyle \sum_{i=1}^{N}{\left({1 + |\gamma_{\theta}\theta_i|}\right)}^{-1}}
\end{equation}
\noindent where $\epsilon_x>0$ is a small constant removing the singularity at $x_i = 0$.

\subsubsection{Total reward}
In order to discourage the agent from simply standing still at a safe location, which would yield a reward of zero given the preliminary reward function, we impose a constant living penalty $r_{exists} < 0$ to the overall reward function. A simple way of setting this parameter is to assume that, given a total absence of nearby obstacles and perfect vessel alignment with the path, the agent should receive a zero reward when moving at a certain slow speed $\alpha_r U_{max}$, where $\alpha_r \in (0, 1)$ is a constant parameter. This gives us 
\begin{equation}
\begin{aligned}
    &r_{exists} + \lambda\left(\left(\tfrac{\alpha_r U_{max}}{U_{max}} + 1\right)\left(1 + 1\right)-1\right) = 0 \\
    &r_{exists} = -\lambda (2 \alpha_r + 1)
\end{aligned}
\end{equation}
\noindent Also, in the interest of having bounded rewards, we enforce a lower bound activated upon collisions by defining the total reward
\begin{equation}
r(t) = \begin{cases}
    \left( 1 - \lambda \right) r_{collision} & \text{\scriptsize (if collision)}\\\
    \lambda r_{pf}(t) + \left( 1 - \lambda \right) r_{oa}(t) + r_{exists} & \text{\scriptsize (otherwise)}
  \end{cases}
\end{equation}
Deciding the optimal value for the trade-off parameter $\lambda$ is a nontrivial endeavour. This touches upon the fundamental challenge tackled in this project, namely how to avoid obstacles while without deviating unnecessarily from the desired trajectory. Thus, we initialize it randomly at each reset of the environment by sampling it from a probability distribution. In order to familiarize the agent with different degrees of radical collision avoidance strategies ($\lambda \to 0$), which is useful in dead-end scenarios where the correct behavior is to ignore the desire for path adherence in order to escape the situation, we sample $\log_{10}{\lambda}$ from 
\begin{equation}\label{eq:lambda_sampling}
    -\log_{10}{\lambda} \sim \textit{Gamma}(\alpha_{\lambda}, \beta_{\lambda})
\end{equation}
\begin{figure}[pos=!htb]
  \centering
    \includegraphics[width=\linewidth]{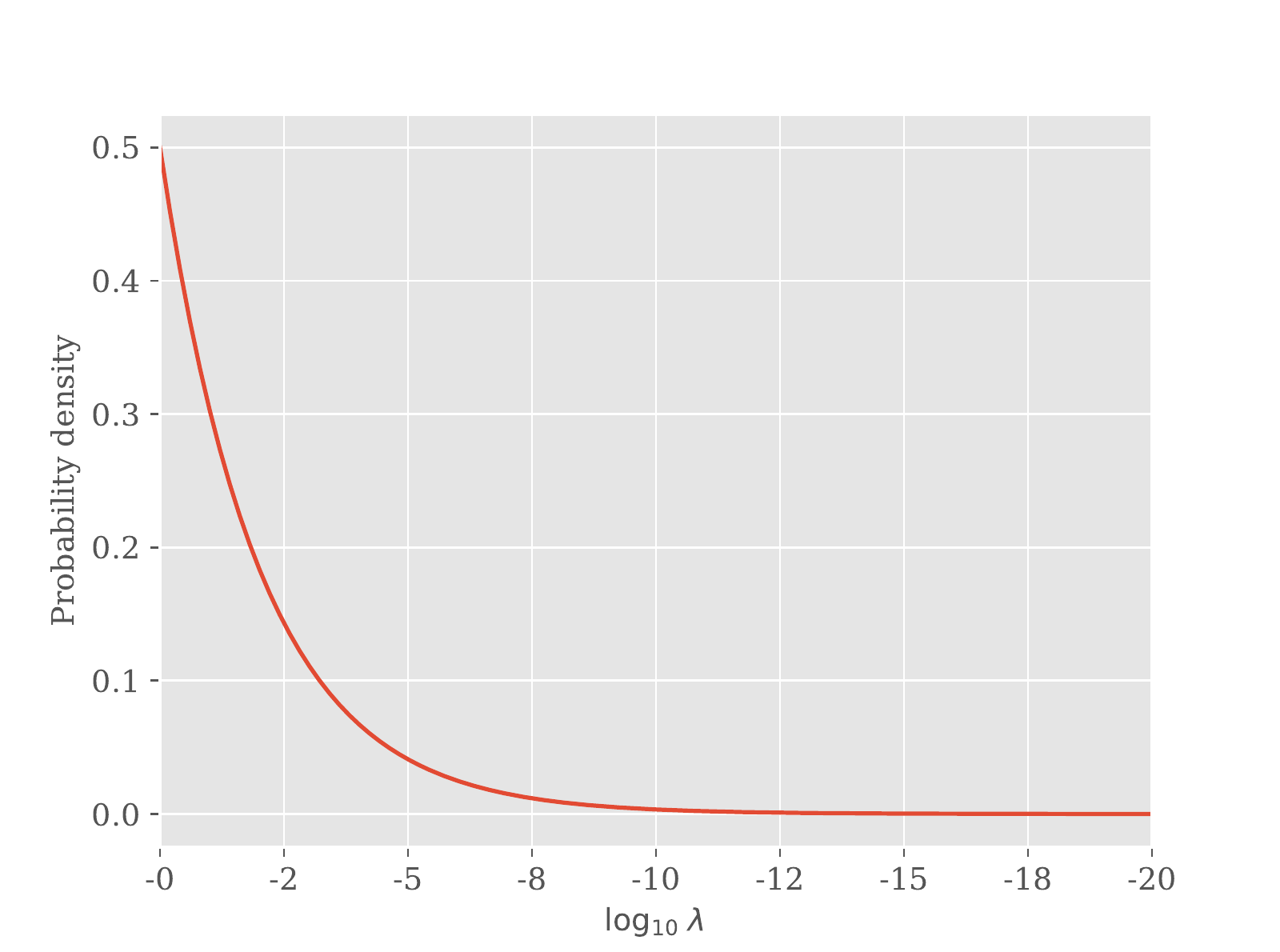}
 \caption[Sampling distribution for trade-off parameter]{Gamma-distribution with parameters $\alpha_{\lambda} = 1$, $\beta_{\lambda} = 2$ from which $-\log_{10}{\lambda} $ is drawn.}
\label{fig:lambda_distr}
\end{figure}
\noindent In order to let the agent base its guidance strategy on the current $\lambda$, we include $\log_{10}{\lambda}$ as an additional observation feature. The reward paremeters used in the current work is given by $\alpha_{\lambda}=1.0$, $\beta_{\lambda}=2.0$, $\gamma_e=0.05$, $\gamma_{\theta}=4.0$, $\gamma_x=0.005$,  $\epsilon_x=1.0m$, $\alpha_r=0.1$, $r_{collision}=-2000$.  


\subsection{Training}

The complete observation vector, which in the context of RL represents the state $s$, contains features representing the position and orientation of the vessel with regards to the path as well as the pooled sensor readings and the logarithm of the current trade-off parameter $\lambda$. 

\begin{table}[pos=h]
\scriptsize
	\begin{tabular}{ll}
		\hline
		\textbf{Observation feature} & \textbf{Definition}\\
		\hline
		Surge velocity & $u^{(t)}$ \\
		Sway velocity & $v^{(t)}$ \\
		Yaw rate & $r^{(t)}$ \\
		Look-ahead course error & 
		$\gamma_p(\bar{\omega}^{(t)} + \Delta_{LA}) - \chi^{(t)}$ \\
		Course error & $\tilde{\chi}^{(t)}$ \\
		Cross-track error & $e^{(t)}$ \\
		Reward trade-off parameter & $\log_{10}{\lambda^{(t)}}$ \\
		Obstacle closeness, first sector & $1 - \frac{1}{S_r}{\text{FeasibilityPooling}(\bm{x}=\{x_1, \dots x_d\})}$ \\
		\vdots \\ Obstacle closeness, last sector &
		$1- \frac{1}{{S_r}} {\text{FeasibilityPooling}(\bm{x}=\{x_{N-d}, \dots x_N\})}$ \\
		\hline
	\end{tabular}
	\caption[Observation vector]{Observation vector $s$ at timestep $t$.}\label{tab:obs_vector}
\end{table}

\noindent The RL agent is trained using the PPO algorithm (ref. Algorithm \ref{alg:ppo}) implemented in the Python library \textit{Stable Baselines} \cite{stable-baselines}, with the hyperparameters given by $\gamma=0.999$, $T=1024$, $N_A=8$, $K=10^6$, $\eta=0.0002$, $N_{MB}=32$, $\lambda=0.95$, $c_1=0.5$, $c_2=0.01$, $\epsilon=0.2$. The action and value function networks were implemented as fully-connected neural networks, both using the $\text{tanh}(.)$ activation function and consisting of with two hidden layers with 64 nodes. We simulate the vessel dynamics using the fifth order Runge-Kutta-Fahlberg method \cite{Fehlberg1970} using the timestep $h=0.14s$. Whenever the vessel either reaches the goal $\bm{p}_{end}$, collides with an obstacle or reaches a cumulative negative reward exceeding $-5000$, the environment is reset according to Algorithm \ref{alg:scenario_creation}.

\subsection{Evaluation}

We analyze the agent's performance based on quantitative as well as qualitative testing. The relationship between the reward trade-off parameter $\lambda$, which is fed to the agent as an observation feature, and the guidance behavior is of particular interest. Specifically, the agent was tested with a range of values from radical path adherence (i.e. $\lambda = 1$) to radical obstacle avoidance ($\lambda \to 0$). The results are listed in Table \ref{tab:test_results_quantitative}.


\subsubsection{Quantitative testing}

In order to obtain statistically significant evidence for the guidance ability of the trained agent, we simulate the agent's behavior in $100$ random environments generated stochastically according to Algorithm \ref{alg:scenario_creation}. We then report the performance criteria in terms of success rate, average cross-track error and average episode length. In the current context, the success rate is defined as the percentage of episodes in which the agent reached goal, average cross-track error is defined as the average deviation from path in meter, average episode length is the average length of episode in seconds. 


\subsection{Qualitative testing}
In addition to the statistical evaluation, we observe the agents' behavior in the test scenarios shown in Figure \ref{fig:test_scenario_sims}.


\section{Results and Discussions}
In this chapter, we present the test results obtained from training and testing the agent and discuss the findings.

\subsection{Training process}

We train the agent for 1157 episodes, corresponding to more than 4 million simulated time-steps of length $\Delta t = 0.14 s$. At this point, all the metrics used for monitoring the training progress had stabilized. The training process, which, for the purpose of faster convergence, ran 8 parallel simulation environments, took approximately 48 hours on a Intel Core i7-8550U CPU.

\subsection{Test results}

As outlined, each value of $\lambda$ was tested for 100 episodes, all of which took place in a randomly generated path following environments according to Algorithm \ref{alg:scenario_creation}. Of course, a larger sample size is always better for quantitative evaluation, but in the interest of time, 100 test episodes for each $\lambda$ value was a reasonable compromise. Clearly, the calculation of the interception points between the rangefinder rays and the obstacles is the most computationally expensive part of the simulation. Thus, the simulation can be made orders of magnitude faster by lowering the sampling rate of the sensors, but we decided to perform the testing without any restrictions to the sensor suite. The observed test results are displayed in Table \ref{tab:test_results_quantitative}.

\begin{table*}[pos=h]
	\begin{tabular}{clccccc}
		\hline
		Agent & \multicolumn{1}{c}{$\lambda$} & Success Rate & \multicolumn{2}{l}{Avg.\,Cross-track Error} & \multicolumn{2}{l}{Avg.\,Episode Length} \\
		\hline
		1 & 1 & 87\% && 22.12 m && 325.0 s \\
	    2 & 0.9 & 84\% && 22.20 m && 324.4 s \\
		3 & 0.5 & 88\% && 23.22 m && 340.8 s \\
	    4 & 0.1 & 95\% && 27.0 m && 343.3 s \\
		5 & 0.01 & 98\% && 33.7 m && 359.1 s \\
	    6 & 0.001 & 99\% && 41.4 m && 386.8 s \\
		7 & 0.0001 & 99\% && 54.8 m && 387.4 s \\
	    8 & 0.00001 & 100\% && 66.8 m && 401.0 s \\
		9 & 0.000001 & 100\% && 80.8 m && 386.9 s \\
		\hline
	\end{tabular}
	\caption[Quantitative test results]{Quantitative test results obtained from 100 episode simulations per agent.}
	\label{tab:test_results_quantitative}
\end{table*}

\begin{figure}[pos=!htb]
    \centering
    \begin{subfigure}{0.49\linewidth}
        \includegraphics[trim={0.5cm 0.1cm 1cm 1cm}, clip, width=\linewidth]{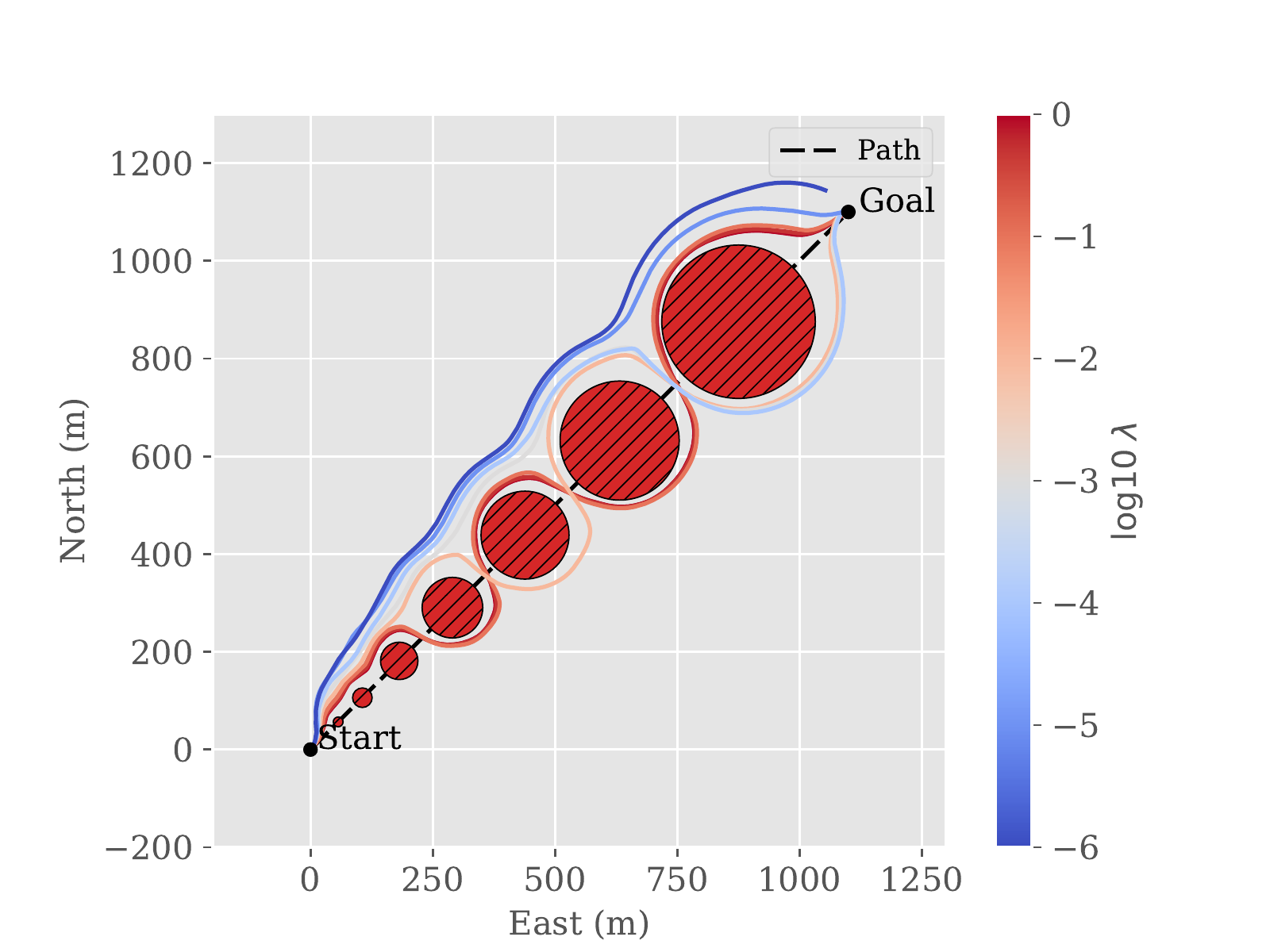}
        \subcaption{Test scenario A}
    \end{subfigure}
    \begin{subfigure}{0.49\linewidth}
        \includegraphics[trim={0.5cm 0.1cm 1cm 1cm},clip,width=\linewidth]{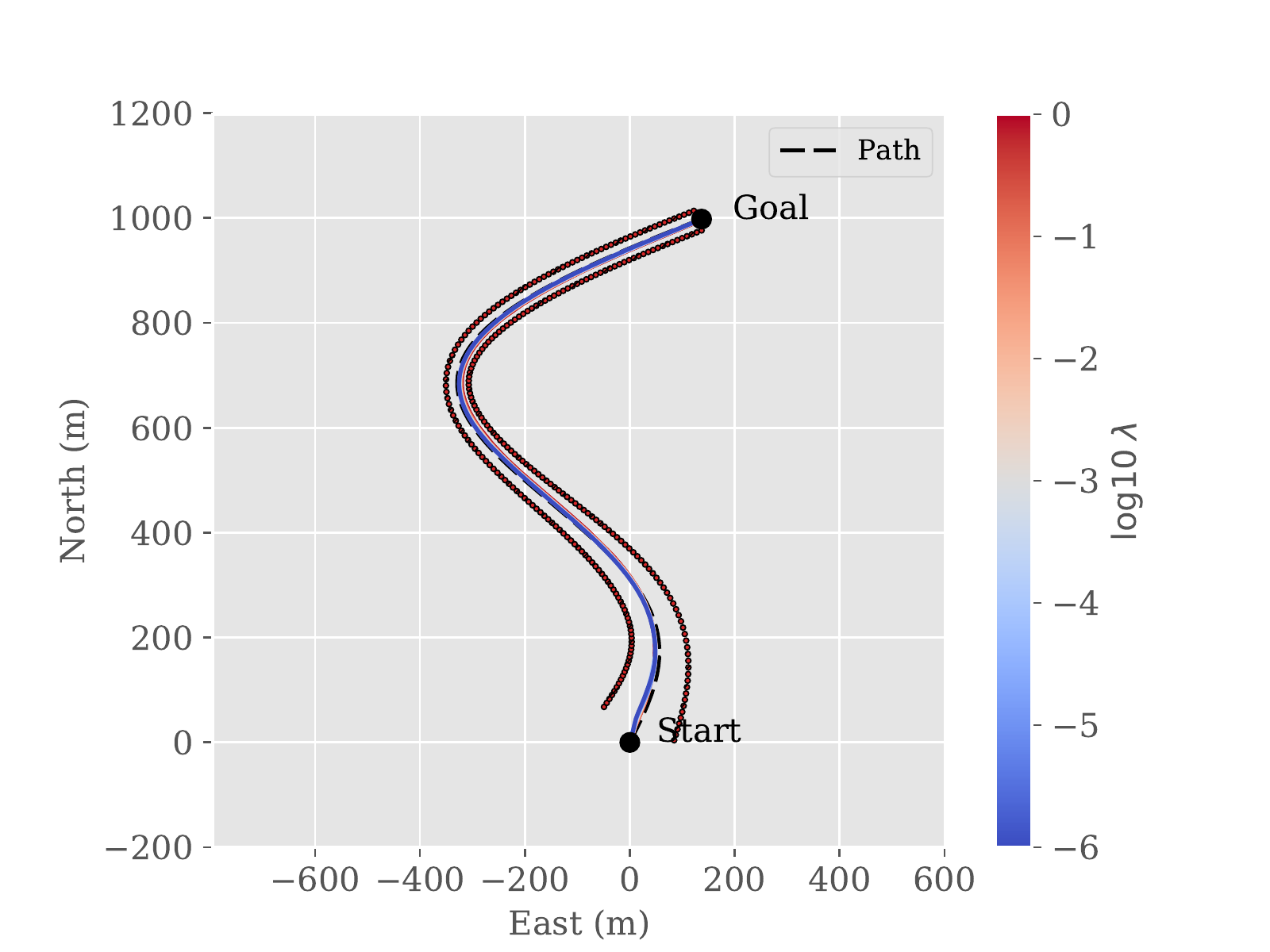}
        \subcaption{Test scenario B}
    \end{subfigure}
    \begin{subfigure}{0.49\linewidth}
        \includegraphics[trim={0.5cm 0.1cm 1cm 1cm},clip,width=\linewidth]{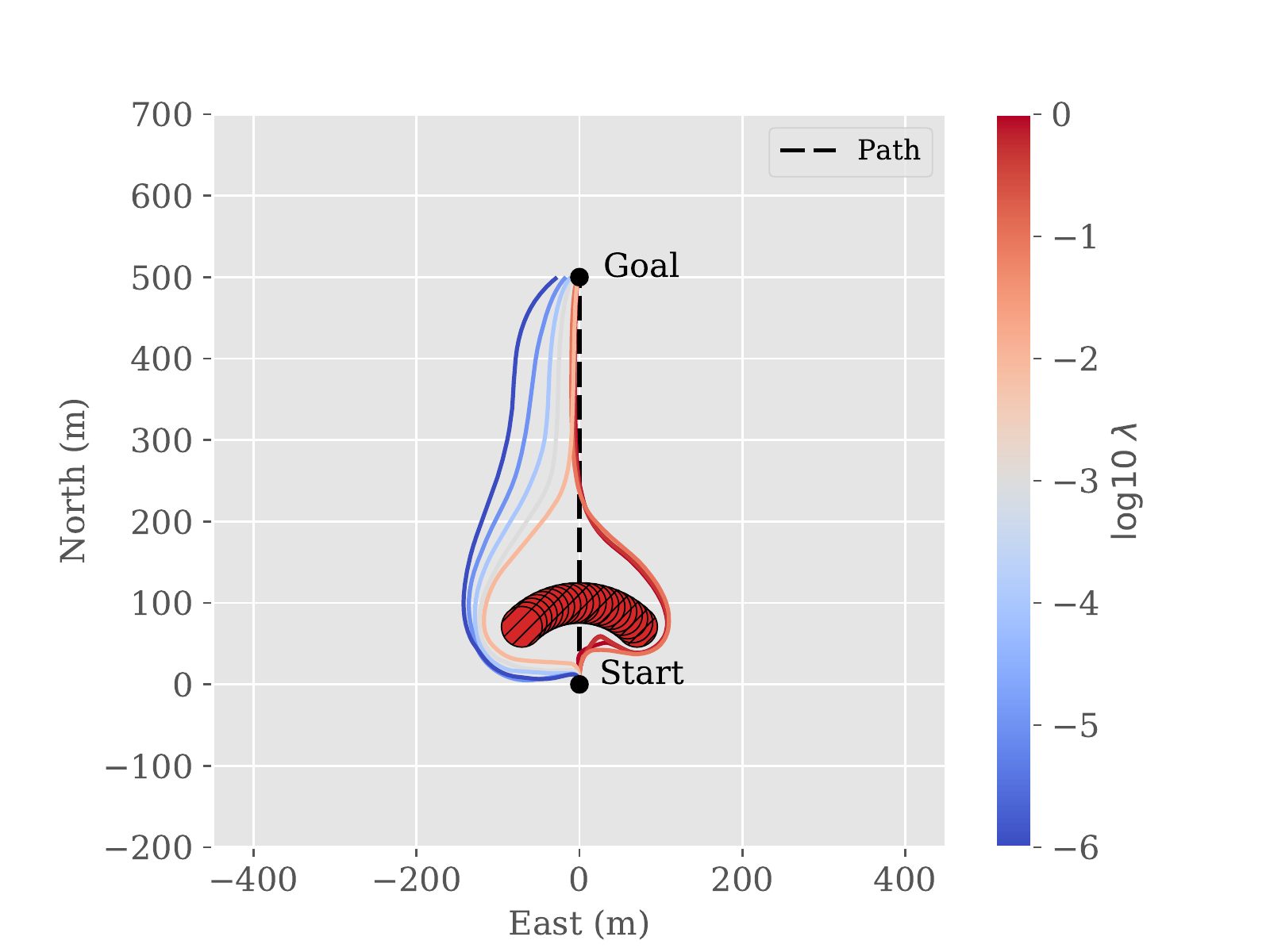}
        \subcaption{Test scenario C}
    \end{subfigure}
    \begin{subfigure}{0.49\linewidth}
        \includegraphics[trim={0.5cm 0.1cm 1cm 1cm},clip,width=\linewidth]{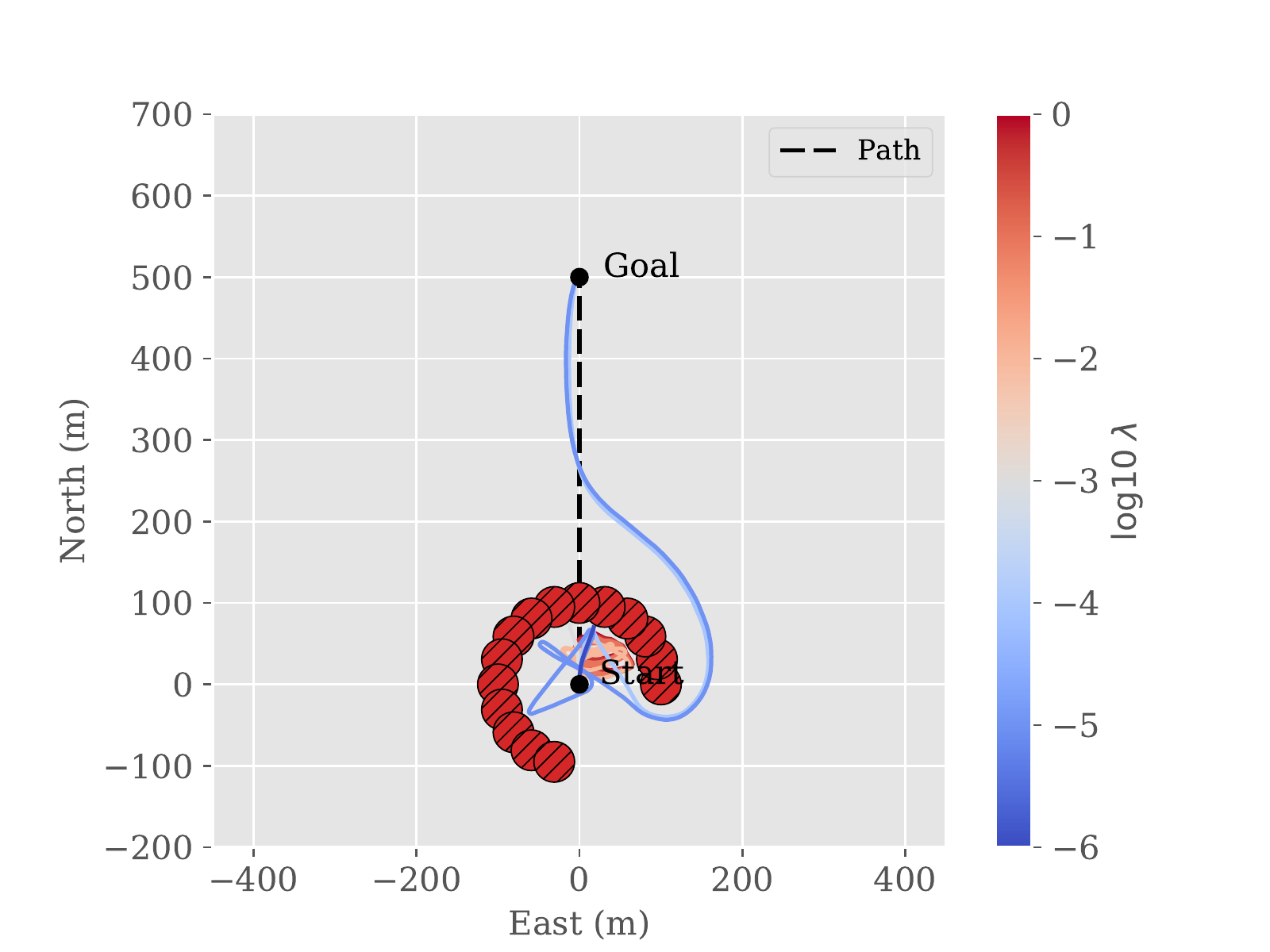}
        \subcaption{Test scenario D}
    \end{subfigure}

	\caption[Test scenario simulations]{Agent trajectories in qualitative test scenarios when $\lambda$ parameter is varied. The behaviour in terms of collision avoidance is significantly modulated.}
	\label{fig:test_scenario_sims}
\end{figure}

Additionally, we simulated each agent in the four outlined qualitative test scenarios. Except for scenario B, in which all agents chose more or less exactly the same trajectory, the other scenarios clearly reflect the differences between the agents. The agents' trajectories in each test scenario are plotted in Figure \ref{fig:test_scenario_sims}.

\begin{figure}[pos=h]
    \centering
    \begin{subfigure}{0.54\linewidth}
        {\includegraphics[height=5.1cm, trim={2.9cm 0 4.4cm 1.3cm}, clip]{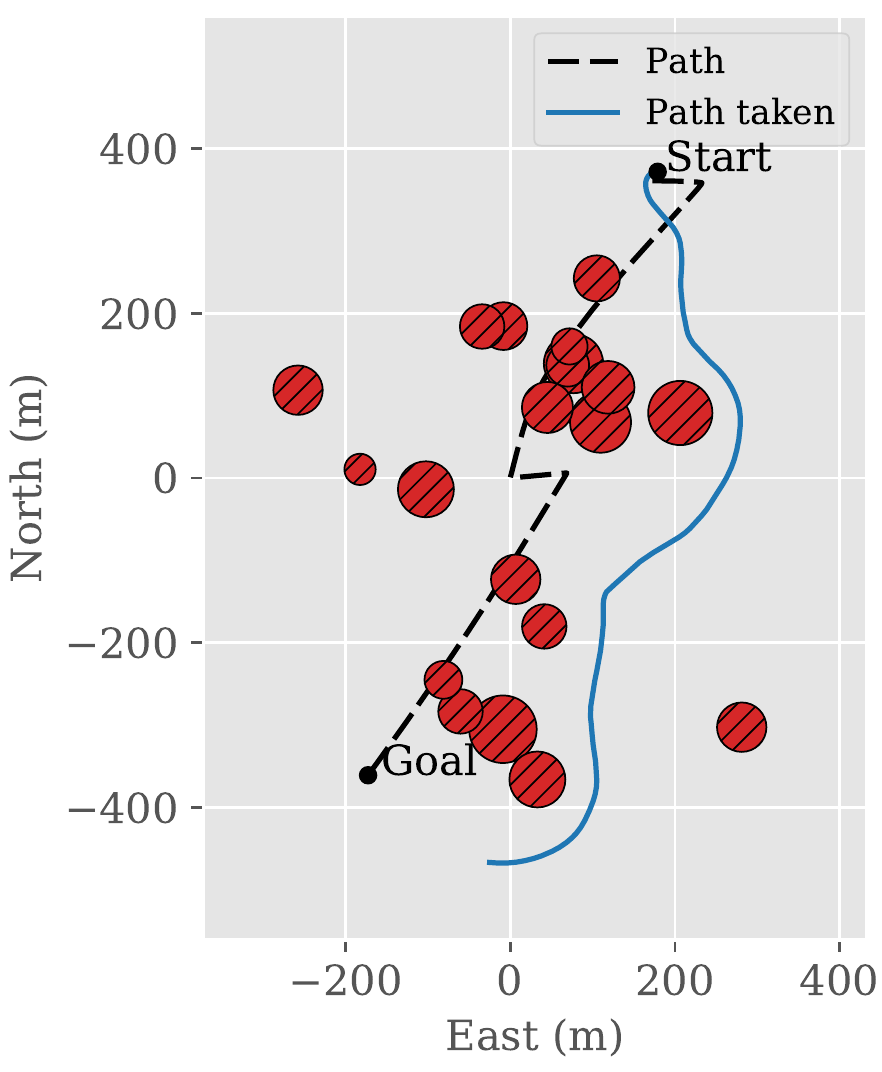}}
        \subcaption{Agent 9, episode 91/100}
        \label{subfig:trajlamsmall}
    \end{subfigure}
    \begin{subfigure}{0.44\linewidth}
        {\includegraphics[height=5.1cm, trim={3.7cm 0 5.1cm 1.3cm}, clip]{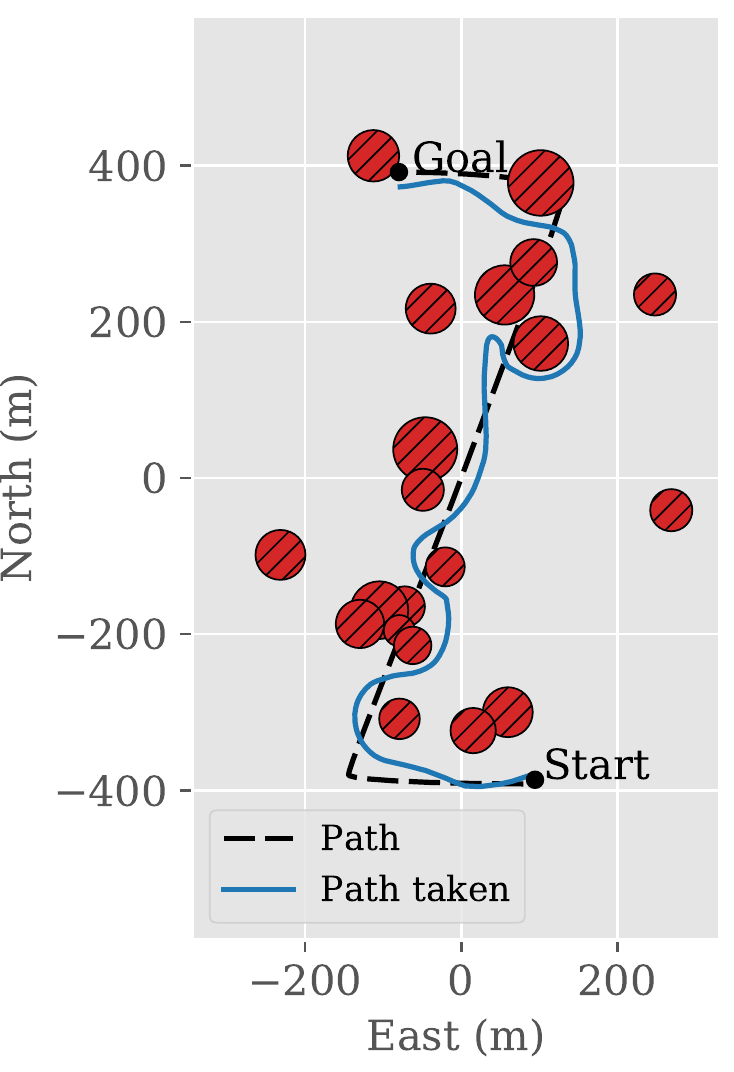}}
        \subcaption{Agent 1, episode 77/100}
        \label{subfig:trajlamlarge}
    \end{subfigure}
	\caption[Example trajectories for extreme trade-off parameter values]{Example trajectories highlighting the difference in guidance strategies for extreme values of the trade-off parameter $\lambda$. Evidently, the radical obstacle avoidance agent, where $\lambda$ was set to $10^{-6}$, clearly exhibits a more defensive behavior, basically avoiding the entire cluster of obstacles surrounding the path \protect\subref{subfig:trajlamsmall}. More impressively, the radical path adherence agent, with $\lambda = 1$, follows the path closely while avoiding the obstacles blocking it \protect\subref{subfig:trajlamlarge}.}
	\label{fig:quantitative_selected_episodes}
\end{figure}

Based on the results, it seems clear that a reactive RL agent is capable of becoming proficient at the combined path-following / collision-avoidance task after being trained using the state-of-the-art PPO algorithm. Prior to conducting any experiments, our assumption was the decreasing $\lambda$, and thus decreasing the degree to which the agent would prioritize path-adherence over collision avoidance, would lead to a higher success rate. Also, our expectation was that this performance increase would come at the expense of the agent's path following performance, leading to an increase in the average cross-track error. The results show a clear and reliable trend, supporting our hypothesis. In fact, as seen in Table \ref{tab:test_results_quantitative}, the collision avoidance rate stabilizes at 100\% when $\lambda$ is sufficiently small. Figure \ref{fig:quantitative_selected_episodes}, which features two episodes extracted from the training process, clearly illustrates why a small $\lambda$ will lead to a lower collision rate, but also cause a significant worsening in path following performance.

\begin{figure}[pos=h]
    \begin{subfigure}{\linewidth}
        \includegraphics[clip,width=\linewidth]{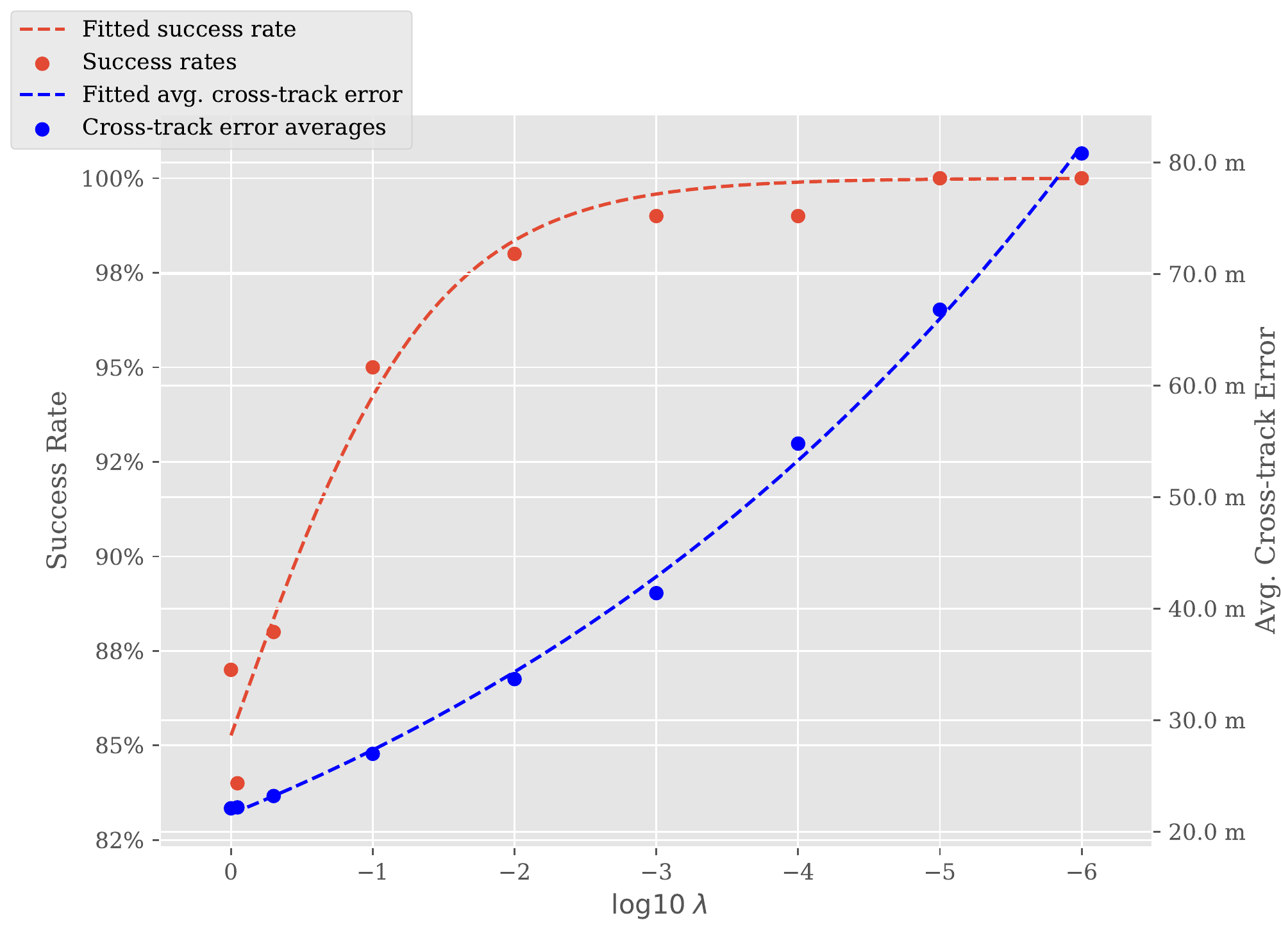}
        \subcaption{The agents empirical success rates and avg. cross-track errors fitted to $\hat{f}(\lambda) = a + \frac{1-a}{1+\lambda^b}$ and $\hat{f}(\lambda) = a + b \lambda^{-c}$, respectively. The non-linear least squares estimate for the success rate model parameters is $a = 0.705,\; b = 0.614$, whereas the estimate for the average cross-track error model parameters is $a = -4.44,\; b = 26.1,\; c = 0.086$}
        \label{fig:successrate_fit}
    \end{subfigure}
	\begin{subfigure}{\linewidth}
        \includegraphics[clip,width=\linewidth]{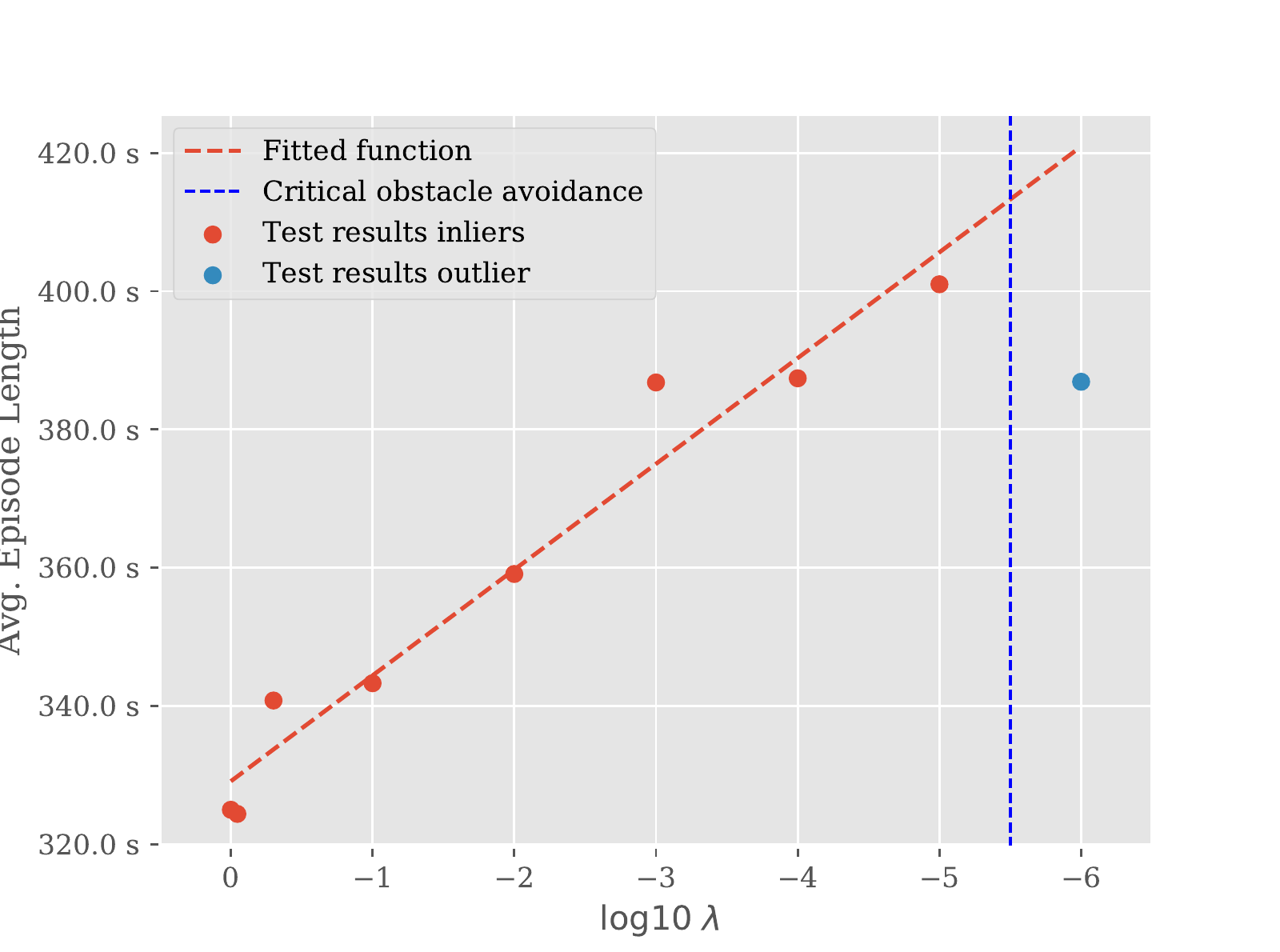}
        \subcaption{The agent's empirical average episode length fitted to $\hat{f}(\lambda) = a - b \log_{10}{\lambda}$. The non-linear least squares estimate for the model parameters is $a = 329,\; b = 15.3$. The point marked as an outlier was excluded from the regression, as there is an obvious explanation as to what might cause a drop in average episode length when $\lambda$ gets very small: due to the resulting radical collision avoidance strategy, the agent will tend to simply avoid the entire cluster of obstacles, instead of avoiding individual obstacles. Thus, the log-linear model will only be valid up to a certain point. In the figure, this validity threshold is labelled as the critical obstacle avoidance}
        \label{fig:episodelength_fit}
    \end{subfigure}
	\caption{Empirical success rate}
\end{figure}

\noindent From plotting the test metrics against $\lambda$, it becomes clear that the trends can be described mathematically by simple parametric functions of $\lambda$. After deciding on suitable parameterizations, we use the Levenberg-Marquardt curve-fit method provided by Python library SciPy \cite{SCIPY} in order to obtain a non-linear least squares estimate for the model parameters. The fitted models for our evaluation metrics can be visualized in Figure \ref{fig:successrate_fit} and Figure \ref{fig:episodelength_fit}. The fitted parametric models allow us to generalize the observed results to unseen values of $\lambda$.

\section{Conclusion}\label{chap:five}
In this work, we have demonstrated that RL is a viable approach to the challenging dual-objective problem of, without relying on a map, controlling a vessel to follow a path given by a priori known way-points while avoiding obstacles along the way. More specifically, we have shown that the state-of-the-art PPO algorithm converges to a policy that yields intelligent guidance behavior under the presence of non-moving obstacles surrounding and blocking the desired path. 

Engineering the agent's observation vector, as well as the reward function, involved the design and implementation of several novel ideas, including the Feasibility Pooling algorithm for intelligent real-time sensor suite dimensionality reduction. By augmenting the agent's observation vector by the reward trade-off parameter $\lambda$, and thus enabling the agent to adapt to changes in its reward function, we have demonstrated through experiments that the agent is capable of adjusting its guidance strategy (i.e. its preference of path-adherence as opposed to collision avoidance) based on the $\lambda$ value that is fed to its observation vector.

By means of extensive testing, we have observed that, even in challenging test environments with high obstacles densities, the agent's success rate is in the high 80s when $\lambda$ is set such that it induces a strict path adherence bias, and close to 100\% when a more defensive strategy is chosen. 

In the end it is important to confess that the DRL algorithms rely heavily on deep neural networks which learn humongous number of trained parameters, interpreting which is humanly impossible currently and is considered a bottleneck in a wholehearted acceptance of these algorithms in safety critical applications. However, the current work do demonstrate the possibility of programming intelligence into these safety critical applications. 
\section*{Acknowledgment}
The authors acknowledge the financial support from the Norwegian Research Council and the industrial partners DNV GL, Kongsberg and Maritime Robotics of the Autosit project. (Grant No.: 295033).
\bibliographystyle{cas-model2-names}
\bibliography{cas-refs}

\bio{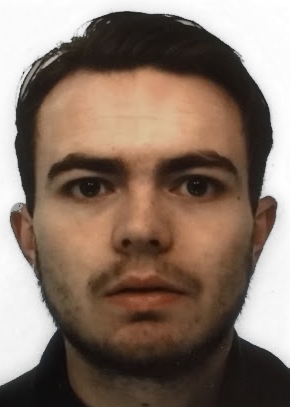}
{Eivind Meyer} is currently working on his Master's thesis, completing his five-year integrated Master's degree in Cybernetics and Robotics at the Norwegian University of Science and Technology (NTNU) in Trondheim. Having specialized in Real Time Systems, his research interests focus
on adopting state-of-the-art Artificial Intelligence methods for Autonomous Vehicle Control.
\endbio
\vskip3pt
\bio{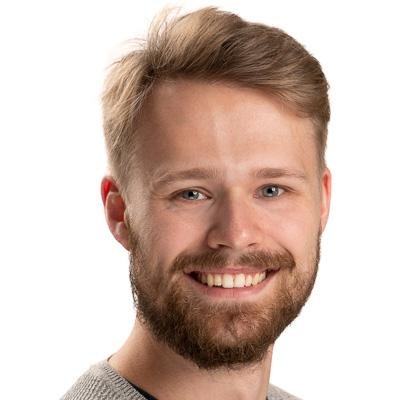}
{Haakon Robinson} is a PhD candidate at the Norwegian University of Science and Technology (NTNU). He received a Bachelors degree in Physics in 2015 and completed a Masters degree in Cybernetics and Robotics in 2019, both at NTNU. His current work investigates the overlap between modern machine learning techniques and established methods within modelling and control, with a focus on improving the interpretability and behavioural guarantees of hybrid models that combine first principle models and data-driven components. 
\endbio
\vskip3pt
\bio{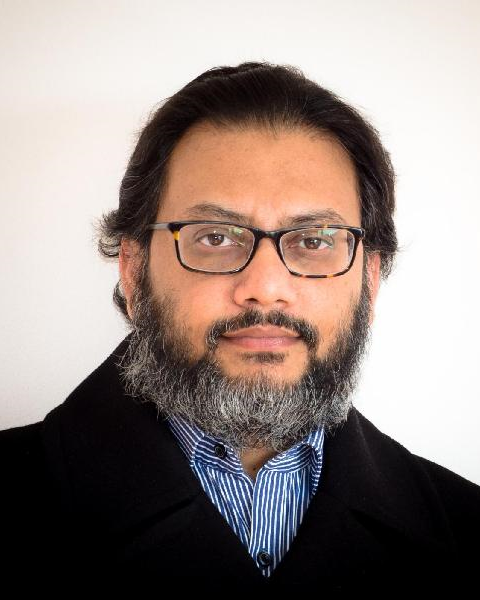}
Adil Rasheed is the professor of Big Data Cybernetics in the Department of Engineering Cybernetics at the Norwegian University of Science and Technology where he is working to develop novel hybrid methods at the intersection of big data, physics driven modelling and data driven modelling in the context of real time automation and control. He also holds a part time senior scientist position in the Department of Mathematics and Cybernetics at SINTEF Digital where he led the Computational Sciences and Engineering group between 2012-2018. He holds a PhD in Multiscale Modeling of Urban Climate from the Swiss Federal Institute of Technology Lausanne. Prior to that he received his bachelors in Mechanical Engineering and a masters in Thermal and Fluids Engineering from the Indian Institute of Technology Bombay.
\endbio
\vskip3pt
\bio{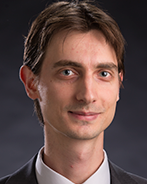}
Omer San received his bachelors in aeronautical engineering from Istanbul Technical University in 2005, his masters in aerospace engineering from Old Dominion University in 2007, and his Ph.D. in engineering mechanics from Virginia Tech in 2012. He worked as a postdoc at Virginia Tech from 2012-'14, and then from 2014-'15 at the University of Notre Dame, Indiana.  
He has been an assistant professor of mechanical and aerospace engineering at Oklahoma State University, Stillwater, OK, USA, since 2015. He is a recipient of U.S. Department of Energy 2018 Early Career Research Program Award in Applied Mathematics. His field of study is centered upon the development, analysis and application of advanced computational methods in science and engineering with a particular emphasis on fluid dynamics across a variety of spatial and temporal scales. 
\endbio

\end{document}